\documentclass[9pt,twocolumn,twoside]{pnas-new}

\templatetype{pnasresearcharticle} 

\usepackage{hyperref}

\def\x{{\mathbf{x}}}

\def\w{{\mathbf{w}}}

\def\dt{\eta}
\def\I{{\mathbf{I}}}
\def\R{{\mathbb{R}}}
\def\E{{\mathbb{E}}}
\def\Ba{{\mathbb{B}}}

\DeclareMathOperator{\Tr}{Tr}

\def\unf{n}
\def\O{\mathcal{O}}
\def\o{o}
\def\vel{\mathbf{v}}
\newcommand{\beq}{\begin{equation}}
\newcommand{\eeq}{\end{equation}}
\newcommand{\beqn}{\begin{eqnarray}}
\newcommand{\eeqn}{\end{eqnarray}}
\newcommand{\lpa}{\left(}
\newcommand{\rpa}{\right)}
\newcommand{\lsq}{\left[}
\newcommand{\rsq}{\right]}
\newcommand{\loss}{l}
\newcommand{\ma}{\kappa}

\newcommand{\erf}{\textrm{erf}}
\newcommand{\ampli}{\lambda}
\newcommand{\aw}{r}
\newcommand{\cov}{\mathbf{\Sigma}}
\newcommand{\ewt}{\mathbf{e}_{1}}
\newcommand{\ewp}{\mathbf{e}_{w_{\perp}}}
\newcommand{\eb}{\mathbf{e}_{\beta}}
\newcommand{\tcross}{{\hat{t}}}

\newcommand{\ec}{\gamma}
\newcommand{\sign}{\mathrm{sign}}

\newcommand{\Bc}{B^*}
\newcommand{\tf}{{t^*}}
\newcommand{\Ste}{S_{test}}

\usetikzlibrary{positioning, shapes, calc, decorations.pathreplacing}
\usepackage{bm}

\def\expModels{\protect\hyperlink{app:expModels}{SI Appendix 1}}
\def\expModelsPercep{\protect\hyperlink{app:expModelsPercep}{SI Appendix 1.A}}
\def\extensions{\protect\hyperlink{app:extensions}{SI Appendix 2}}
\def\extensionsWD{\protect\hyperlink{app:extensionsWD}{SI Appendix 2.B}}
\def\nnPlots{\protect\hyperlink{app:dnnExp}{SI Appendix 3}}
\def\stochastic{\protect\hyperlink{app:stoch_terms}{SI Appendix 4.B}}
\def\heuristic{\protect\hyperlink{app:break}{SI Appendix 5}}
\def\sdeAverages{\protect\hyperlink{app:summary}{SI Appendix 6.A}}
\def\sdeDerivation{\protect\hyperlink{app:onlineSDE_w1}{SI Appendix 6.B}}

\def\figNTK{\ref{fig:ntk} }
\def\figStripe{\ref{fig:stripe} }
\def\figFCphase{\ref{fig:FCfeature} }
\def\figFCphaseES{\ref{fig:FCfeature}-(d) }
\def\figCNNphase{\ref{fig:CNNfeature} }
\def\figCNNphaseES{\ref{fig:CNNfeature}-(d) }
\def\figAlign{\ref{fig:FC_dynamics_feature}-(a) }
\def\figBcrit{\ref{fig:FC_dynamics_feature}-(b) }
\def\figCNNbatch{\ref{fig:CNN_batch}-(a) }
\def\figTcross{\ref{fig:early_stopping} }
\def\figTP{\ref{fig:n_w_t} }
\def\figTPw{\ref{fig:wnorm} }
\def\figWmargin{\ref{fig:wnorm}-(b)}

\setboolean{displaywatermark}{false}
\begin{document}

\title{On the different regimes of stochastic gradient descent}

\author[a,1]{Antonio Sclocchi}
\author[a,1]{Matthieu Wyart}

\affil[a]{Institute of Physics, Ecole Polytechnique F\'ed\' erale de Lausanne, 1015 Lausanne, Switzerland}

\leadauthor{Sclocchi}

\significancestatement{The success of deep learning contrasts with its limited understanding. One example is stochastic gradient descent, the main algorithm used to train neural networks. It depends on hyperparameters 
whose choice has little theoretical guidance and relies on expensive trial-and-error procedures. 
In this work, we clarify how these hyperparameters affect the training dynamics of neural networks, leading to a phase diagram with distinct dynamical regimes. Our results explain the surprising observation that these hyperparameters strongly depend on the number of data available. We show that this dependence is controlled by the difficulty of the task being learned.}

\authordeclaration{}
\correspondingauthor{\textsuperscript{1}To whom correspondence should be addressed. E-mail: antonio.sclocchi@epfl.ch, matthieu.wyart@epfl.ch}


\begin{abstract}
Modern deep networks are trained with stochastic gradient descent (SGD) whose key hyperparameters are the number of data considered at each step or batch size $B$, and the step size or learning rate $\dt$.
For small $B$ and large $\dt$, SGD corresponds to a stochastic evolution of the parameters, whose noise amplitude is governed by the ``temperature'' $T\equiv \dt/B$. Yet this description is observed to break down for sufficiently large batches $B\geq \Bc$, or simplifies to gradient descent (GD) when the temperature is sufficiently small. Understanding where these cross-overs take place remains a central challenge. Here, we resolve these questions for a teacher-student perceptron classification model and show empirically that our key predictions still apply to deep networks. Specifically, we obtain a phase diagram in the $B$-$\eta$ plane that separates three dynamical phases: \textit{(i)} a noise-dominated SGD governed by temperature, \textit{(ii)} a large-first-step-dominated SGD and \textit{(iii)} GD.
These different phases also correspond to different regimes of generalization error. 
Remarkably, our analysis reveals that the batch size $\Bc$ separating regimes \textit{(i)} and \textit{(ii)} scale with the size $P$ of the training set, with an exponent that characterizes the hardness of the classification problem. 
\end{abstract}


\maketitle
\thispagestyle{firststyle}
\ifthenelse{\boolean{shortarticle}}{\ifthenelse{\boolean{singlecolumn}}{\abscontentformatted}{\abscontent}}{}

\begin{figure*}
    \centering
    \includegraphics[width=.84\textwidth]{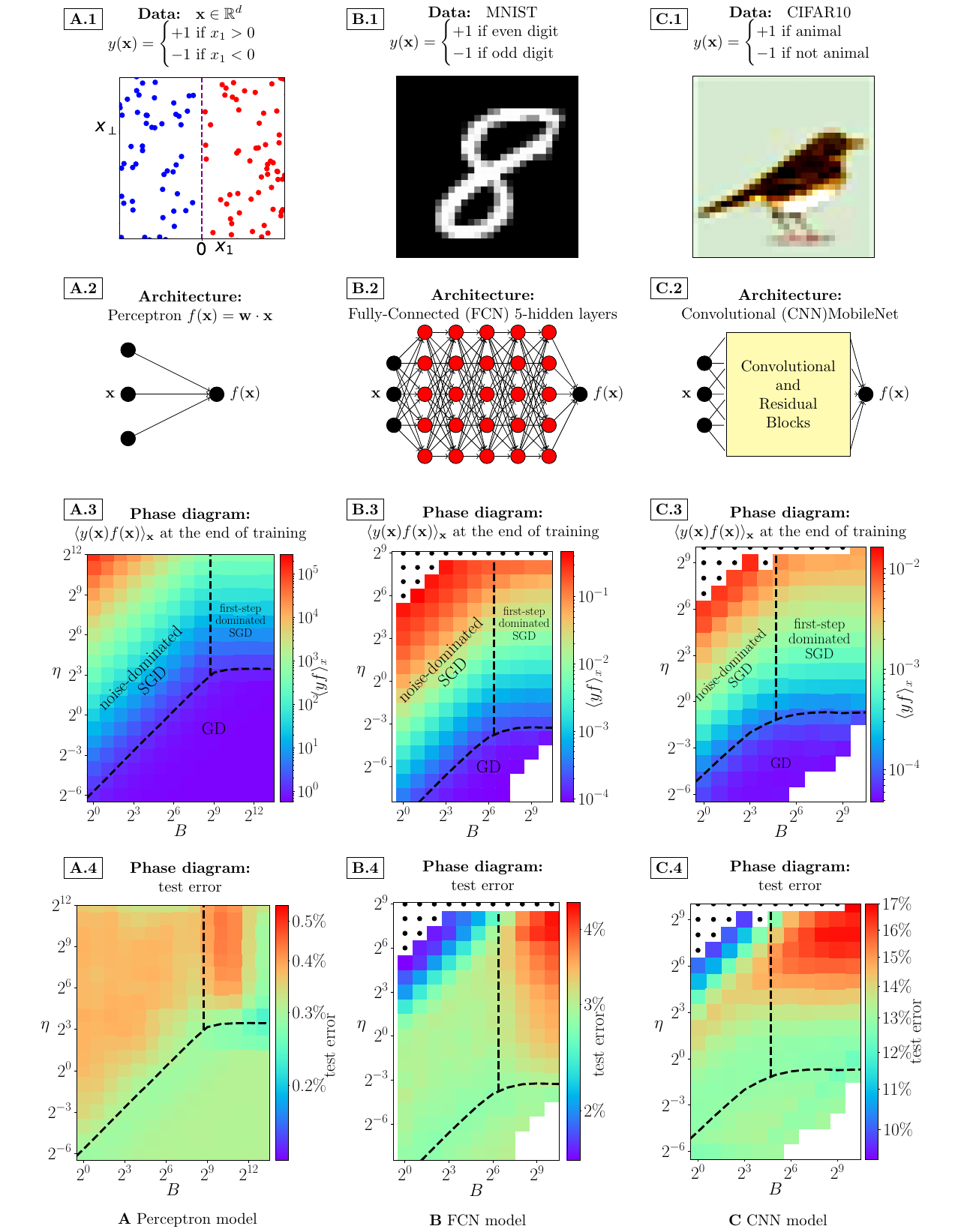}
    \caption{\textbf{ SGD phase diagrams for different data and architectures.}
    The different data sets considered are a teacher perceptron model with $P=8192$, dimension $d=128$ and data distribution of Eq. \ref{eq:rho_x1} with $\chi=1$ \textit{(A.1)}, $P=32768$ images of  MNIST \textit{(B.1)} and $P=16384$ images of CIFAR 10 \textit{(C.1)}. The different neural network architectures trained on these datasets correspond respectively to \textit{(A.2)} a perceptron, for which the output is linear in both the input $\x$ and the weights $\w$, and trained with hinge loss margin $\ma=2^{-7}$;
    \textit{(B.2)} a fully-connected network  with $5$ hidden layers,  $128$ hidden neurons per layer and margin $\ma=2^{-15}$;
    \textit{(C.2)} a CNN  made by several blocks composed of depth-wise, point-wise and standard convolutions plus residual connections (more details in \expModels), with margin $\ma=2^{-15}$.
    Panels \textit{(A.3)},\textit{(B.3)},\textit{(C.3)} display the alignment after training in the $\dt,B$ phase diagram.
    The black dots correspond to diverging trainings where the algorithm does not converge.
    We can distinguish the noise-dominated SGD regime, for which the alignment is constant along the diagonals $\frac{\dt}{B}=T$. Within the first-step-dominated SGD, instead, the alignment is constant at constant $\dt$.  For small $\dt$, one enters in the gradient descent (GD) regime where the alignment does not depend on $\dt$ and $B$.
    Taking this value $m_{GD}$ of the alignment as a reference, the black dashed line $\eta_c(B)$ delimiting the GD region corresponds to the alignment taking value $2 m_{GD}$.
    The vertical black dashed line guides the eye to indicate the critical batch size $\Bc$. 
    Panels \textit{(A.4)},\textit{(B.4)},\textit{(C.4)} display the test error again as a function of $\dt,B$. As expected, the test error is constant along the diagonals $\frac{\dt}{B}=T$ for noise-dominated SGD and constant in the GD regime. For first-step-dominated SGD, the test error can be affected by both $\dt$ and $B$, and can improve at very large batches (see discussion below). 
    }
    \label{fig:phase}
\end{figure*}

{\Firstpage

\dropcap{S}tochastic gradient descent (SGD), with its variations, has been the algorithm of choice to minimize the loss and train neural networks since the introduction of back-propagation \cite{robbins1951stochastic, lecun2002backprop, bottou2010large}. 
When minimizing an empirical loss on a training set of size $P$, SGD consists in estimating the loss gradient using a mini-batch of the data selected randomly at each step. When the size $B$ of the batch is small, SGD becomes a noisy version of gradient descent (GD) and the magnitude of this noise is controlled by the `temperature' $T=\dt/B$, with $\dt$ the learning rate \cite{jastrzkebski2017}.
This temperature scale is not a thermodynamic quantity, since in general no thermal equilibrium is reached before the dynamics.
However, understanding the scale of hyperparameters where this description holds and SGD noise matters has remained a challenge.
Specific questions include \textit{(i)} below which temperature $T_c$ noise becomes irrelevant and the dynamics corresponds to gradient descent?
\textit{(ii)} What determines the critical batch size $\Bc$, beyond which SGD is observed not to be controlled by temperature in a variety of settings \cite{shallue2019, smith2020}? This question is of practical importance: after searching for an optimal temperature, practitioners can maximize the batch size up to $\Bc$ while keeping temperature fixed, as large batches can lead to faster computations in practice \cite{goyal2017accurate, mccandlish2018empirical}.
\textit{(iii)}  It was observed that the variation of the network weights during training increases as power laws of both $T$ and $P$, both for deep nets and for simple models like the perceptron  \cite{dissectingSGD}. Yet, a quantitative explanation of this phenomenon is lacking.

SGD noise has attracted significant research interest, in particular for its connections to loss landscape properties and performance. Recent works describing SGD via a stochastic differential equation (SDE) have emphasized several effects of this noise, including its ability to escape saddles \cite{ge2015escaping, jordan2017saddles, daneshmand2018escaping, hu2019diffusionSGD, zhu2019anisotropic, benarous2022, florent2023unifyingSGD}, to bias the dynamics toward broader minima \cite{zhang2018energy,wu2018dynamical,xie2021diffusion, tu2021, tu2023} or toward sparser distributions of the network parameters \cite{flammarion2021, vivien2022label, ganguli2023, flammarion2023, Yu2023SGD}. Yet, most of these works assume that fresh samples are considered at each time step, and are thus unable to explain the dependence of SGD on the finite size of the training set $P$. Here, we study this dependence on classification problems and we identify three regimes of SGD in deep learning. Following \cite{dissectingSGD}, we first consider a perceptron setting where classes are given by a teacher $y(\x)=\sign(\w^*\cdot\x)$, where $\w^*$ is a unit vector, and are learned by a student $f(\x)\propto \w\cdot\x$. We solve this problem analytically for the hinge loss, which vanishes if all data are fitted by some margin $\kappa$.
Our central results, summarized in Fig. \ref{fig:phase}A and the phase diagrams therein, are as follows.

{\it Noise-dominated SGD.} For small batches and large learning rates, the dynamics is well-described by an SDE with a noise scale $T=\dt/B$.
We show that this noise controls the components $\w_\perp$ of the student weights orthogonal to the teacher, such that at the end of training $\|\w_\perp\|\sim T$.
Using this result, together with considerations on the angle of the predictor that can fit the entire training set, implies that the weight magnitude 

} 
\noindent after training indeed depends both on $T$ and $P$, as $\|\w\|\sim T P^{\gamma}$, thus explaining the observation of \cite{dissectingSGD}. 
The exponent $\gamma$ characterizes the difficulty of the classification problem.
{\it Gradient Descent.} This regime breaks down when the temperature, and therefore the magnitude of orthogonal weights $\|\w_\perp\|$, is too small. Then, fitting all data no longer corresponds to a constraint on the angle of the predictor $\|\w_\perp\|/\|\w\|$, but instead to a constraint on $\|\w\|$ to satisfy the margin. This latter constraint is the most stringent one for  $T\leq T_c\sim \kappa$. In that regime,  temperature can be neglected and the dynamics corresponds to gradient descent, thus answering point \textit{(i)} above. 
  
{\it First-step-dominated SGD.} The noise-dominated SGD regime also breaks down when the batch and learning rates are increased at fixed $T$. Indeed, the first step will increase the weight magnitude to $\|\w\|\sim \eta$, which is larger than the noise-dominated prediction $\|\w\|\sim T P^{\gamma}$ if $B\geq \Bc\equiv P^\gamma$, answering \textit{(ii)} above.

Our second central result is empirical: we find this phase diagram also in deep neural networks.
We train using the hinge loss as in the perceptron case (using the cross-entropy loss with early stopping leads to similar performance and weights increase \cite{spigler2019jamming, dissectingSGD}).
In this case, it is useful to introduce the quantity $\langle y(\x) f(\x) \rangle_{\x}$ characterizing the magnitude with which the output aligns to the task (for the perceptron, it simply corresponds to the quantity $\w\cdot\w^*$).  Fig. \ref{fig:phase} shows this alignment in the diagram $\eta,B$ for fully connected  (B) and Convolutional nets (C), revealing in each case the three regimes also obtained for the perceptron. As shown below, in both cases, $B^*$ depends as a power law of the training set size, as predicted.

\subsection{Related Works}
{\it SGD descriptions.}
In the high-dimensional setting, several works have analyzed online-SGD in teacher-student problems \cite{engel2001statistical, saad1995dynamics, saad1995online, goldt2019dynamics, veiga2022phase}. It has been rigorously shown \cite{benarous2022} that, for online-SGD in high dimensions, the effective dynamics of summary statistics obey an ordinary differential equation with correction terms given by SGD stochasticity.
Our analysis of the perceptron falls into this regime, with fixed points of the dynamics depending explicitly on $T=\dt/B$.
We further introduce the effects of a finite training set $P$ by studying when this online description breaks down.
Other studies \cite{mignacco2020dynamical, mignacco2022effective} have analyzed the correlations of SGD noise with the training set using dynamical mean-field theory. They consider the regime where the batch size $B$ is extensive in the training set size $P$, while our online description considers the regime where $B$ is much smaller than $P$.\\
In finite-dimensional descriptions of SGD, theoretical justification for approximating SGD with an SDE has generally relied upon vanishing learning rates \cite{li2019stochastic, hu2019diffusionSGD}. This condition is too restrictive for standard deep learning settings and \cite{arora2021validity} has shown empirically that the applicability of the SDE approximation can be extended to finite learning rates.
Other SDE approximations assume isotropic and homogeneous noise \cite{welling2011bayesian, neelakantan2015adding, ge2015escaping, raginsky2017non} or a noise covariance proportional to Hessian of the loss \cite{jastrzkebski2017, zhu2019anisotropic, hoffer2017, xie2021diffusion},
while in the online limit of the perceptron model we can compute the exact dependence of the noise covariance on the model weights.
Other studies have observed a link between the effects of SGD noise and the size of the training set \cite{smith2018bayesian}. In our work, we show that these effects do not depend on a direct increase of the noise scale with the training set size, but rather on the fact a larger $P$ implies that larger weights are needed to fit the data, a process that depends on SGD noise.\\
{\it Critical batch size.}
An empirical study at fixed training set size of large-batch training has observed in several learning settings that $\Bc$ is inversely proportional to the signal-to-noise ratio of loss gradients across training samples \cite{mccandlish2018empirical}.  
Our work is consistent with these findings, but most importantly further predicts and tests a non-trivial power law dependence of $\Bc$ with $P$.

\section{Noise-Dominated Regime}
\label{sec:definition}

\subsection{Perceptron Model}

We consider data $\x \in \R^d$ with $d\gg 1$ that are linearly separable. Without loss of generality, we chose that the class label $y(\x)=\pm 1$ is given by the sign of the first component $y(\x) = \textrm{sign}(x_{1})$.
$\{\mathbf{e}_i\}_{i=1,...,d}$ is the canonical basis and $\ewt$ corresponds to the teacher direction. The informative component $x_1$ of each datum is independently sampled from the probability distribution
\beq
\rho(x_1) = |x_1|^\chi e^{-x_1^2/2} / Z,
\label{eq:rho_x1}
\eeq
where $Z$ is a normalization constant 
and $\chi > -1$ \cite{tomasini2022failure}. The other $d-1$ components $\x_\perp = [x_i]_{i=2,...,d}$ are distributed as standard multivariate Gaussian numbers, i.e. $\x_\perp \sim \mathcal{N}(\mathbf{0}, \I_{d-1})$, being $\I_{d-1}$ the $(d-1)$ identity matrix. The parameter $\chi$ controls the data distribution near the decision boundary $\x_1=0$ and how `hard' the classification problem is. Smaller $\chi$ corresponds to harder problems, as more points lie close to the boundary. The case $\chi=0$ corresponds to the Gaussian case.
\footnote{Although the theory is valid for any $\chi>-1$, estimates of $\chi$ in image datasets correspond to $\chi>0$ \cite{tomasini2022failure} (see discussion in Section \ref{sec:experiments}).}

As an architecture, we chose  the perceptron $f(\w,\x) =  \w \cdot \x/\sqrt{d}$. The weights are trained by minimizing the hinge loss 
$L(\w) = \frac{1}{P} \sum_{\mu=1}^P (\ma-y^{\mu} f(\w,\x^{\mu}))^+$,
where $(x)^+ = \max(0,x)$, $\ma>0$ is the margin, $\{(\x^{\mu},y^\mu=y(\x^\mu))\}_{\mu=1,...,P}$ is the training set with size $P\gg d$. We denote $\w^t$ the weights obtained at time $t$, defined as the number of training steps times the learning rate, starting from the initialization $\w^0=\mathbf{0}$.\\
The hinge loss is minimized with a SGD algorithm, in which the batch is randomly selected at each time step among all the $P$ data. The learning rate $\dt$ is kept constant during training. The end of training is reached at time $\tf$ when $L(\w^{\tf})=\unf(\w^{\tf})=0$, where $\unf(\w^t)$ indicates the fraction  of data not fitted at time $t$:  $\unf(\w) = \frac{1}{P} \sum_{\mu=1}^P \theta(\ma-y^{\mu} f(\w,\x^{\mu}))$,
with $\theta(\dots)$ the Heaviside step function.

\subsection{Stochastic Differential Equation}
\label{sec:dyn_descr}

In the limit of small batches $B\ll P$, SGD can be described with a stochastic differential equation (SDE) \cite{li2019stochastic}:
\beq
d\w^t = - dt \nabla L(\w^t) + \sqrt{T} \sqrt{\cov\lpa \w^t\rpa} d\mathbf{W}^t
\label{eq:general_SDE}
\eeq
where $\mathbf{W}^t$ is a $d$-dimensional Wiener process (Ito's convention) and $\cov\lpa \w\rpa/B$ is the covariance of the mini-batch gradients \cite{chaudhari2018},  given in \expModelsPercep. We first make the  approximation of substituting the empirical gradient $\nabla L(\w^t)$ and this covariance with their population averages
$\nabla L(\w^t)\approx \mathbf{g}^t = \E_{\x}\lsq -\nabla_\w\loss(\w^t,\x)\rsq$ and $\cov^t \approx \E_{\x}\lsq \nabla_\w\loss(\w^t,\x) \otimes \nabla_\w\loss(\w^t,\x)\rsq - {\mathbf{g}^t}\otimes {\mathbf{g}^t}$. This approximation does not include finite training set effects. Our strategy below is to estimate the time where such a simplified description breaks down; the solution of the SDE at this time provides the correct asymptotic behaviors for the network at the end of training, as we observe experimentally.

\subsubsection{Asymptotic online-SDE dynamics}

We decompose the student weights as $\w = w_1\ewt + \w_\perp$ and we study the dynamics of both the scalar variables $w_1^t$ (the `overlap' between the student and teacher \cite{engel2001statistical}), and the magnitude of the weights in the perpendicular directions $\|\w_\perp^t\|$.  
Given that we consider a data distribution that is radially symmetric in $\x_\perp$, but not in $\x$, $w_1$ and $\|\w_\perp\|$ are the natural order parameters in our problem. Therefore, the $d$-dimensional dynamics of $\w$ can be studied through a $2$-dimensional summary statistics.

One finds that the online-SDE dynamics of $w_1^t$ and $\|\w_\perp^t\|$, using Ito's lemma (\sdeDerivation), can be written as:

\begin{align}
\begin{split}
\begin{cases}
    dw_1^t &= dt\ g_1^t + \sqrt{\frac{T}{d}} \sigma_1^t d\tilde{W}_1^t\\
    d\|\w_\perp^t\| &= dt\ \lsq g_\perp^t + \frac{T}{2\|\w_\perp^t\|} 
    \unf^t \rsq
    + \sqrt{\frac{T}{d}} \sigma_2^t d\tilde{W}_2^t
\end{cases}
\end{split}
\label{eq:dynamics}
\end{align}
where $\tilde{W}_1^t$ and $\tilde{W}_2^t$ are Wiener processes and the expressions for $g_1^t$, $g_\perp^t$, $\unf^t$, $\sigma_{1}^t$, $\sigma_{2}^t$, reported in \sdeAverages, are functions of $w_1^t$ and $\|\w_\perp^t\|$ through the time-dependent ratios:
\beq
\ampli=\frac{w_1^t}{\|\w_\perp^t\|},\quad
\aw = \frac{\ma\sqrt{d}}{\|\w_\perp^t\|}.
\eeq
The quantity $\ampli$ measures the angle $\theta$ between the student and the teacher directions, since $\theta=\arctan{\ampli^{-1}}$.
The ratio $\aw$ compares the hinge loss margin $\ma$ with the magnitude of the orthogonal components $\|\w_\perp^t\|$.

In the limit $d\gg 1$, the stochastic part of Eq. \ref{eq:dynamics} is negligible, as we show in \stochastic.
The variables $w_1^t$ and $\|\w_\perp^t\|$, therefore, have a deterministic time evolution given by the deterministic part of Eq. \ref{eq:dynamics}.
This result has been proved more generally for the summary statistics of different learning tasks in high dimensions \cite{benarous2022}. Note that, even if the stochastic fluctuations are negligible, the SGD noise affects the dynamics through the term $\frac{T}{2\|\w_\perp^t\|}n^t$ in the evolution of $\|\w_\perp^t\|$.

The term $g_1^t$ (\sdeAverages) is always positive and vanishes in the limit $\ampli\rightarrow \infty$, which determines the fixed point of Eq. \ref{eq:dynamics}. Therefore, we consider its vicinity
\beq
    \ampli=\frac{w_1^t}{\|\w_\perp^t\|}\rightarrow\infty,
    \label{eq:asy_lim}
\eeq
corresponding to a vanishing angle $\theta=\arctan{\ampli^{-1}}\sim \ampli^{-1}$ between the student and teacher directions. In addition, we consider the limit 
\beq
    \aw = \frac{\ma\sqrt{d}}{\|\w_\perp^t\|}\rightarrow 0,
    \label{eq:asy_lim1}
\eeq
and argue below why this limit corresponds to the noise-dominated regime of SGD.

Under the conditions of Eqs. \ref{eq:asy_lim}-\ref{eq:asy_lim1}, the deterministic part of Eq. \ref{eq:dynamics} reads:
\begin{align}
    \begin{split}
        \begin{cases}
        dw_1^t &= dt\ \ampli^{-\chi-2}\ \frac{c_1}{\sqrt{d}} \\
        d\|\w_\perp^t\| &= dt\ \ampli^{-\chi-1} \lsq-\frac{1}{\sqrt{d} \sqrt{2\pi}}+ \frac{T}{2\|\w_\perp^t\|} c_n\rsq,
        \end{cases}
    \end{split}
    \label{eq:asy_dyn}
\end{align}
with constants $c_1$, $c_n$ (\sdeAverages).
Solving Eq. \ref{eq:asy_dyn} gives:
\beq
\|\w_\perp^t\| \propto T \sqrt{d},\quad
w_1^t \sim T \sqrt{d} \lpa \frac{t}{Td}\rpa^{\frac{1}{3+\chi}}.
\label{eq:asy_sol}
\eeq
Therefore the orthogonal component $\|\w_\perp^t\|$ tends to a steady state proportional to the SGD temperature $T$, while the informative one $w_1^t$ grows as a power law of time.
Eq. \ref{eq:asy_sol} implies that $\ampli\gg 1$ is obtained in the large time limit $t\gg Td$, while $\aw\ll 1$ corresponds to $T\gg\ma$, and therefore holds at sufficiently high temperatures. 

\subsubsection{SDE breakdown and solution obtained at the end of training}
Due to the finiteness of $P$, the online solution of Eq. \ref{eq:asy_sol} is expected to be no longer valid at some time $\tcross<\tf$. 
In \heuristic, we provide an argument, confirmed empirically, predicting that $\tcross$ is reached when the number of training points contributing to the hinge loss gradient is of order $\O(d)$. This is obtained by assuming that the student perceptron follows the dynamics Eq. \ref{eq:asy_sol} and applying the central limit theorem to the empirical gradient $\nabla L(\w^t)$. One finds that the magnitude of its population average follows $\|\E_{\x}\lsq\nabla L(\w^t)\rsq\|\sim \frac{1}{\sqrt{d}}\unf(\w^t)$ while that of its finite-$P$ fluctuations is given by 
$\frac{1}{\sqrt{P}}\sqrt{\unf(\w^t)}$. 
For $P$ finite but much larger than $d$, the average gradient is much larger than its fluctuations as long as the fraction of unfitted training points $\unf(\w^t)$ satisfies $\unf(\w^t)\gg \frac{d}{P}$. 
As the training progresses, more data points get fitted and $\unf(\w^t)$ decreases until reaching the condition
\beq
\unf(\w^\tcross)\sim\O\lpa\frac{d}{P}\rpa.
\label{eq:arg_tcross}
\eeq
After this point, the empirical gradient and the population one greatly differ, and the online dynamics is no longer a valid description of the training dynamics. This is the time beyond which test and train errors start to differ.
At leading order in $\ampli\rightarrow\infty$ as $r\rightarrow 0$, we have that $\unf(\w^t) \sim \ampli^{-\chi-1}$. In fact, for a density of points $\rho(x_1)\sim x_1^\chi$ at small $x_1$, $\unf(\w^t)\sim\theta^{\chi+1}\sim \ampli^{-\chi-1}$ is the fraction of points with coordinate $x_1$ smaller than $\theta\sim\ampli^{-1}$.
\footnote[1]{A training point $(\x, y)$ has a non-zero hinge loss and contributes to $\unf(\w)$ if $y f(\x)<\ma$, that is $|x_1|< \frac{\|\w_\perp\|}{w_1} \lpa c + \frac{\ma\sqrt{d}}{\|\w_\perp\|}\rpa$, with $c=-\frac{y \w_\perp\cdot\x_\perp}{\|\w_\perp\|}$, which becomes $|x_1|< c \ampli^{-1}$ for $\aw=\frac{\ma\sqrt{d}}{\|\w_\perp\|}\rightarrow 0$. In the limit $\ampli\rightarrow\infty$, the condition $|x_1|< c\ \ampli^{-1}$ corresponds to the scaling relationship $\unf(\w)\sim \int_0^{\ampli^{-1}}dx_1\rho(x_1)\sim \ampli^{-\chi-1}$, since $\rho(x_1)\sim x_1^\chi$ for $x_1\rightarrow 0$.}

Therefore, using Eq. \ref{eq:asy_sol},
\beq
\unf(\w^t) \sim \ampli^{-\chi-1}
    \sim \lpa \frac{t}{Td}\rpa^{-\frac{\chi+1}{\chi+3}}.
\label{eq:n_asymp}
\eeq
Note that, for $t\ll \tcross$, $\unf(\w^t)$ corresponds to the test error.
Neglecting the dependence in $d$, we finally obtain:
\beq\label{eq:tcross_wcross}
    \tcross \sim 
    T\ P^b,\quad
    \|\w^\tcross\| \sim w_1^\tcross \sim 
    T P^\ec,
\eeq
where $b=1+\frac{2}{1+\chi}$ and $\ec=\frac{1}{1+\chi}$.\\

In Figure \figTcross, we show experimentally that the asymptotic solution of the online SDE is a valid description of SGD up to $\tcross$, as predicted by Eq. \ref{eq:arg_tcross}.
In Figure \figTP-\figTPw, we observe empirically that the power law scaling in $T$ and $P$ of the stopping time $\tf$ and of $\|\w^{\tf}\|$ are the same as those of $\tcross$ and $\|\w^\tcross\|$, therefore  Eqs. \ref{eq:tcross_wcross} also hold to characterize the end of training.

\subsection{T Condition for the Noise-Dominated Regime}
To fit a data point $(\x^\mu, y^\mu)$, the perceptron weights must satisfy the condition $y^\mu \frac{w_1 x_1^\mu}{\sqrt{d}} + y^\mu\frac{\w_\perp\cdot\x_\perp^\mu}{\sqrt{d}} \geq \kappa$, which corresponds to \cite{dissectingSGD}:
\beq
    \frac{w_1}{\|\w_\perp\|} \geq \frac{1}{|x_1^\mu|}\lpa\frac{\ma\sqrt{d}}{\|\w_\perp\|}+c^\mu \rpa,
    \label{eq:fit}
\eeq
where we have defined the random quantity $c^\mu  = -y^\mu\frac{\w_\perp}{\|\w_\perp\|}\cdot\x_\perp^\mu={\cal O}(1)$, and where 
$\frac{\ma \sqrt{d}}{\|\w_\perp\|}\propto \frac{\ma}{T}$ in the noise-dominated regime where Eq. \ref{eq:asy_sol} applies.

For $T\gg\ma$, $\frac{\ma \sqrt{d}}{\|\w_\perp\|}$ is negligible with respect to $c^\mu={\cal O}(1)$ and Eq. \ref{eq:fit} now becomes $\frac{w_1}{\|\w_\perp\|}\geq \frac{c^\mu}{|x_1^\mu|}$, independent of the margin $\ma$. In this case, therefore, fitting $(\x^\mu, y^\mu)$ is constrained by the SGD noise which inflates the non-informative weight component $\|\w_\perp\|$.
For $T\ll\ma$, instead, fitting $(\x^\mu, y^\mu)$ is constrained by the margin $\ma$ and the SGD noise is negligible in Eq. \ref{eq:fit}, 
implying that the temperature delimiting the noise-dominated regime of SGD follows:
\beq
T_c\propto \ma.
\eeq
For $T\ll T_c$, the final magnitude of $\|\w_\perp\|$ is independent of $T$ as expected. Instead, one finds that it is proportional to $\ma$ (Fig. \figWmargin).

\section{Critical Batch Size $\Bc$}
\label{sec:Bcrit}

\begin{figure}
    \centering
    \includegraphics[width=1\columnwidth]{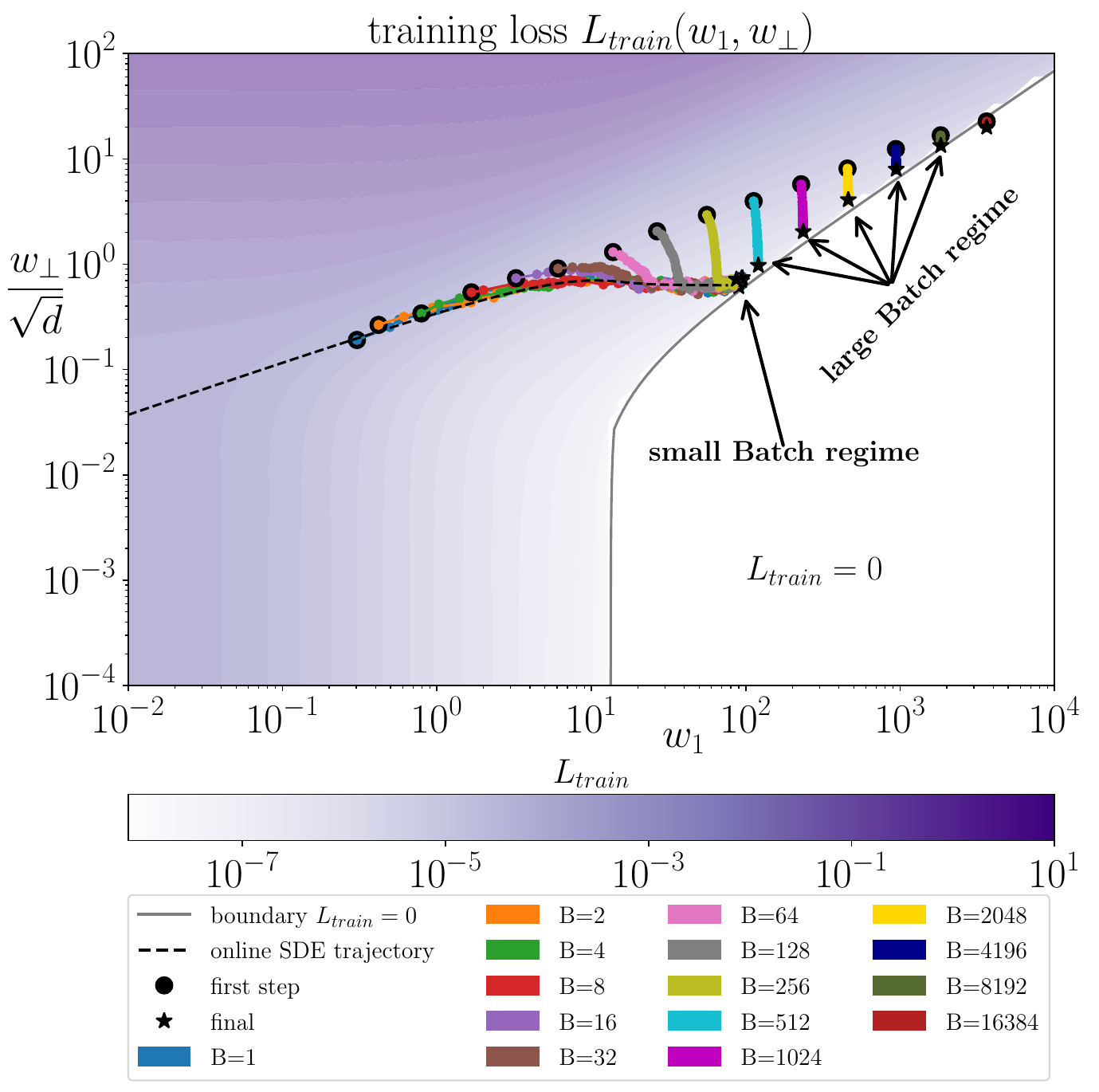}
    \caption{
     \textbf{Dynamical trajectories of SGD in the $(w_1,w_\perp)$ plane, at fixed $T$ and varying batch size as indicated in caption.} Black circles indicate the first step of SGD, black stars indicate the last one. For small enough batches (and therefore small learning rates), trajectories converge to the online SDE solution (black dashed line). For large batches, this is not true anymore, and the final magnitude of the weights increases with batch size.  The location of stopping weights corresponds to zero loss, which can be approximately determined by measuring the hinge loss values $L_{train}(w_1,w_\perp)$ (shown in color) computed as a function of the perceptron weights $\w = w_1\ewt + w_\perp \bm{\xi}$. Here, $\bm{\xi}$ is a $(d-1)$-dimensional Gaussian random vector.
    The white area corresponds to interpolating solutions $L_{train}=0$ in this simplified set-up.     
    For full-batch, we observe that $\w$ can land directly in the white area and therefore fit the data with at most few steps. This behavior affects the test error when $\dt$ is large (Fig. \ref{fig:phase}-A4). 
    Data correspond to $P=16384$, $d=128$, $\ma=0.01$, $\chi=1$, $T=2$.
    }
    \label{fig:trainloss}
\end{figure}

In the small batch regime $B\ll \Bc$, for a given temperature $T=\eta/B$, different SGD simulations converge to the asymptotic trajectory given by the online-SDE, as observed in Fig. \ref{fig:trainloss}  showing SGD trajectories in the $(w_1,w_\perp)$ plane, at fixed $T$ and varying batch size.
This is not the case in the large batch regime $B\gg\Bc$. In fact, for fixed $T$, a larger batch size corresponds to a larger learning rate and therefore a larger initial step. If the initial step is too large, SGD would converge to a zero-loss solution before being able to converge to the online-SDE asymptotic trajectory. This situation is vividly apparent in Fig. \ref{fig:trainloss}.
Therefore, we can estimate the scale of the critical batch-size $\Bc$  by comparing the first step $ w_1^\dt \propto \dt$ with the final value $w_1^{\tf}\sim \frac{\dt}{B}P^\ec$ in the noise-dominated regime.
When $w_1^\dt\ll w_1^{\tf}$, SGD converges to the asymptotic online-SDE and the final minimum depends on $T=\dt/B$.
When $w_1^\dt\gg w_1^{\tf}$, SGD does not converge to the asymptotic online-SDE and the final weight magnitude depends only on $w_1^\dt$, as shown in Fig. \ref{fig:m_dynamics}.
The condition $w_1^\dt\sim w_1^{\tf}$, that is $\dt \sim \frac{\dt}{\Bc} P^\ec$ gives:
\beq
    \Bc \sim P^\ec,
    \label{eq:BcP}
\eeq
with $\ec=\frac{1}{1+\chi}$ depending on the data distribution. The relationship of Eq. \ref{eq:BcP} is well verified from the empirical data of Fig. \ref{fig:Bcrit}-(a).\\
This finding is consistent with $\Bc$ being inversely proportional to the signal-to-noise ratio of loss gradients across training samples as proposed in \cite{mccandlish2018empirical}. In fact, in our setting, increasing the gradient noise scale by a factor $\sigma$ corresponds to the substitution $T\rightarrow \sigma T$, which transforms Eq. \ref{eq:BcP} into $\Bc \sim \sigma P^\ec$. Moreover, at large $P$, increasing $\chi$ reduces the exponent $\ec$ and $\Bc$. In fact, larger $\chi$ corresponds to fewer training points close to the decision boundary and therefore a larger gradients' signal-to-noise ratio.

\subsection{Performance in Large Step SGD}
The first training step leads to $w_1^\dt = \frac{\dt}{B} \sum_{\mu\in\Ba_1} \frac{|x_1^\mu|}{\sqrt{d}}$ and $\w_\perp^\dt = \frac{\dt}{B} \sum_{\mu\in\Ba_1} y^\mu\frac{\x_\perp^\mu}{\sqrt{d}}$, where $\Ba_1$ is the first sampled batch. 
$\w_\perp^\dt$ is a zero-mean random vector of norm $\|\w_\perp^\dt\|=\O( \frac{\dt}{\sqrt{B}})$,
while $w_1^\dt=\O( \dt)$. We can distinguish several regimes:
\begin{itemize}
    \item[(i)] If $w_1^\dt|x_1^\mu| \sim \dt |x_1^\mu|\gg \kappa\ \forall \mu$, then the margin $\kappa$ can be neglected. From extreme value statistics $\min_\mu |x_1^\mu|=\O(P^{-1/(1+\chi)})$, thus this condition is saturated when $\dt\sim \kappa P^{1/(1+\chi)}$. It corresponds to the horizontal dashed line in the diagrams of Fig. \ref{fig:phase}.
    \item[(ii)]  Above this line in the regime $\dt\gg \kappa P^{1/(1+\chi)}$, the fraction of unfitted points as well as the test error after one step are given by the angle of the predictor $\lambda^\dt= \frac{w_1^\dt}{\|\w_\perp^\dt\|}\sim\O(\sqrt{B})$. Following Eq. \ref{eq:n_asymp},  $\unf(\w^\eta) \sim (\lambda^\dt)^{-\chi-1}\sim B^{-(\chi+1)/2}$. If $\unf(\w^\eta)\ll 1/P$ or equivalently $B\gg P^{2/(1+\chi)}$, then with high probability all points are fitted after one step and the dynamics stops; the final test error is then of order $\epsilon\sim B^{-(\chi+1)/2}$. Since $B\leq P$, this condition can only occur for $\chi\geq 1$.  Note that the error is smaller than in the noise-dominated regime where $\epsilon\sim 1/P$. 
\end{itemize}
For the marginal case $\chi=1$ in Fig. \ref{fig:trainloss}, we observe that the full batch case $B=16384$ reaches (nearly) zero training loss after the first training step. Correspondingly, in the phase diagram of Fig. \ref{fig:phase}-A4,  the lowest test error is achieved for full batch in the first-step-dominated regime.

\subsection{Notes on Variations of SGD}
Our analysis does not consider variations of SGD sometimes used in practice, such as the addition of momentum, adaptive learning rates, or weight decay. A detailed extension of the present approach to these cases is an interesting work left for the future. In  \extensions, we provide evidence supporting that adding momentum is akin to rescaling the learning rate by a momentum-dependent factor. By contrast, adding weight decay in the noise-dominated regime is akin to stopping the dynamics early, and it is therefore lowering performance in the case of the perceptron with no label noise. For large batch sizes, the first-step-dominated dynamics eventually converge to the noise-dominated one at a time-scale depending on the weight-decay coefficient (see \extensionsWD).

\section{Experiments for Deep Networks}
\label{sec:experiments}

\begin{figure}
    \centering
    \includegraphics[width=1\columnwidth]{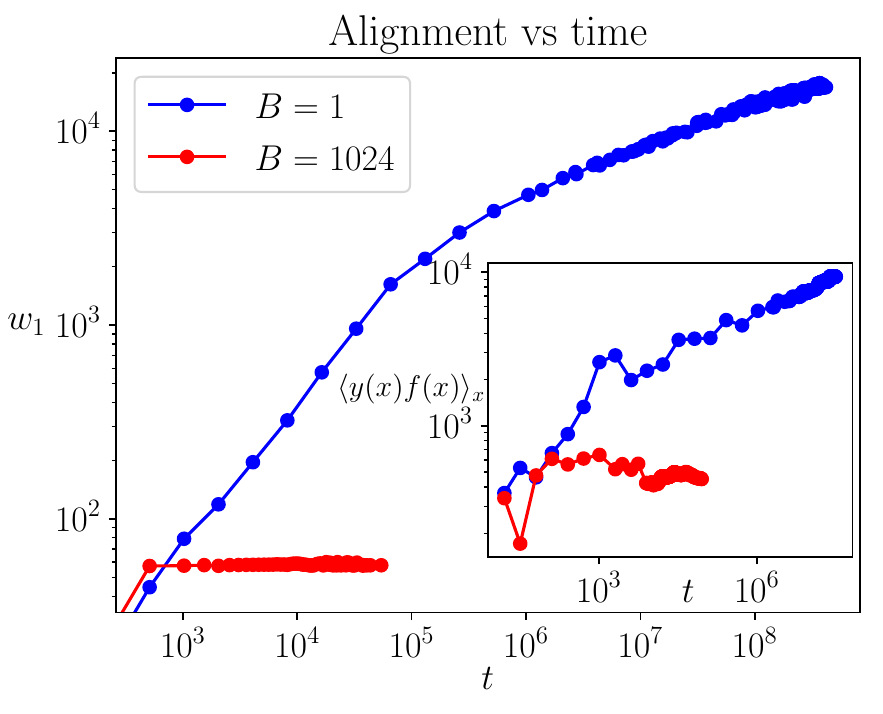}
    \caption{\textbf{For large learning rates, the dynamics of the alignment is different in the small batch and large batch regimes.} (Main panel) Perceptron: in this case, the alignment is proportional to the student component $w_1$; data for fixed $\dt=512$, same setting as Fig. \ref{fig:phase}. For small $B$, $w_1$ grows during the training dynamics, while, for large $B$, its final value is reached after a single training step. (Inset) Fully connected network on MNIST, small margin ($\ma=2^{-15}$), fixed $\dt=16$, same setting as Fig. \ref{fig:phase}: For small and large batch, the alignment shows a similar dynamics to the perceptron case, although for large batch it reaches its final value after some training steps (and not just a single step).
    }
    \label{fig:m_dynamics}
\end{figure}

\begin{figure*}
    \centering
    \includegraphics[width=.95\textwidth]{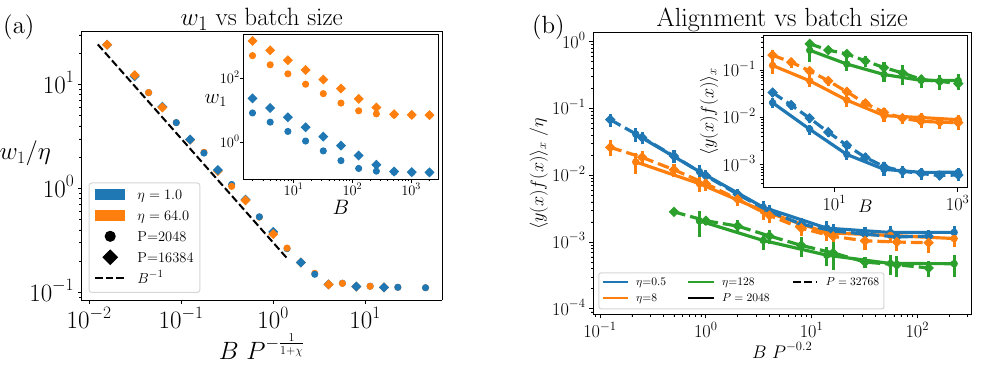}
    \caption{
    \textbf{The critical batch size $\Bc$ depends on the size of the training set as predicted by Eq. \ref{eq:BcP}.}
    (a)  $w_1$ at the end of training for the perceptron. Inset: $w_1$ depends on $\dt$, $B$ and $P$ for small $B$, while it only depends on $\dt$ for large $B$. We observe that the cross-over between small and large $B$ depends on $P$.
    Main panel: the curves collapse in one curve by rescaling $w_1$ by $\dt$ and $B$ by $\Bc\propto P^\ec$, $\ec=\frac{1}{1+\chi}$. This is consistent with $w_1\sim\frac{\dt}{B}P^\ec$ for small $B$ (Eq. \ref{eq:tcross_wcross}) and $w_1\sim \dt$ for large $B$ (section \ref{sec:Bcrit}).
    (b) Fully-connected network on parity MNIST. Alignment $\langle y(\x) f(\x)\rangle_\x$ at the end of training (measured as in Eq. \ref{eq:align}). Inset: as for $w_1$ in panel (a), $\langle y(\x) f(\x)\rangle_\x$ depends on $\dt$, $B$ and $P$ for small $B$, and it only depends on $\dt$ for large $B$.
    Main panel: rescaling $B$ by $P^{0.2}$ aligns the cross-over batch size at different $P$s, suggesting a dependence $\Bc\propto P^{0.2}$. The curves are approximately collapsed by the rescaling of the y-axis as $\langle y(\x) f(\x)\rangle_\x/\dt$.
    }
    \label{fig:Bcrit}
\end{figure*}

We consider a $5$-hidden-layer fully-connected architecture with GELU activation functions \cite{gelu} that classifies the parity MNIST dataset (even vs. odd digits) and a deep convolutional architecture (MobileNet) which classifies the CIFAR10 images (animals vs. the rest). As before, training corresponds to minimizing the  hinge loss with SGD at constant $\dt$, until reaching zero training loss, see \expModels\ 
for further details.

For deep neural networks,  the alignment between the network output $f(\x)$ and the data labels $y(\x) \in \{+1,-1\}$ cannot simply be written as a projection $w_1$ of the weights in some direction. Instead, we define it as:
\beq
\langle y(\x) f(\x) \rangle_\x = \frac{1}{|\Ste|}\sum_{\x^\nu\in \Ste} y(\x^{\nu}) f(\x^\nu).
\label{eq:align}
\eeq
where the average is made over the test set $\Ste$. For the perceptron, $w_1$ and $\langle y(\x) f(\x) \rangle_\x$ are proportional.

\subsection{Small Margin $\kappa$}

An interesting parameter to vary is the margin $\kappa$, which is equivalent to changing the initialization scale of the network. For very small $\kappa$ and gradient flow, tiny changes of weights can fit the data, corresponding to the lazy regime where neural nets behave as kernel methods \cite{jacot2018, chizat2019lazy, geiger2020disentangling}. However, by cranking up the learning rate one escapes this regime \cite{catapult2020}, which is the case in our set-up (Fig. \figNTK in \nnPlots). Our central results  are as follows:
\begin{itemize}
    \item[(i)] The phase diagram predicted for the perceptron holds true, as shown Fig. \ref{fig:phase}-B3,C3 representing the alignment in the ($B$, $\dt$) plane.  We observe a noise-dominated SGD where the alignment depends on the ratio $T=\dt/B$; a first-step-dominated SGD where the alignment is constant at a given $\dt$; and a  GD regime at small $\dt$ where the alignment does not depend on $\dt$ and $B$.
    \item[(ii)] These three regimes also delineate different behaviors of the test error, as shown in Fig. \ref{fig:phase}-B4,C4.
    \item[(iii)] Fig. \ref{fig:m_dynamics}-\textit{inset} confirms the prediction that in the first-step-dominated SGD regime, the alignment builds up in a few steps. By contrast, in the noise-dominated SGD regime, it builds up slowly over time.
    \item[(iv)] The critical batch $\Bc$ separating these two regimes indeed depends on a power law of the training set  $P$.
    This result is revealed in Fig. \ref{fig:Bcrit}-(b),  reporting the alignment at the end of training as a function of the batch size $B$. The inset shows that the alignment depends on the batch size for small $B$, while it only depends on $\dt$ for large $B$. The cross-over $\Bc$ between the two regimes depends on $P$. It is estimated in the main panel by rescaling the x-axis by $P^{-0.2}$ so as to align these cross-overs, indicating a dependence $\Bc \sim P^{0.2}$. The same phenomenology is observed for the perceptron in Fig. \ref{fig:Bcrit}-(a), with the relationship $\Bc\sim P^{\frac{1}{1+\chi}}$ given by Eq. \ref{eq:BcP}.
    We can use this measurement of $\Bc$ in neural networks to estimate a value of $\chi$ for the corresponding dataset. From the data of Fig. \ref{fig:Bcrit}-(b), we obtain $\chi_{MNIST}\approx 4$ for the MNIST dataset, while from Fig. \figCNNbatch $\chi_{CIFAR}\approx 1.5$ for CIFAR10. These estimates are relatively close to the values $6$ and $1.5$, for MNIST and CIFAR10 respectively, independently measured in a study of kernel ridge regression \cite{tomasini2022failure}. In that context, the value of $\chi$ directly affects the kernel eigenvalue spectrum and therefore the task difficulty. These estimates suggest that our toy model for $\chi>0$ captures some properties of image classification tasks. 
\end{itemize}

\subsection{Large Margin $\kappa$}

In the alternative case where the margin is large, the predictor has to grow in time or  ``inflate'' in order to become of the order of the margin and fit the data \cite{jacot2018, geiger2020disentangling,jacot2021saddle}. As a result, the weights and the magnitude of the gradients initially grow in time. In our set-up, this regime can be obtained choosing $\kappa=1$, because at initialization the output function is small. \footnote{We chose the NTK initialization \cite{jacot2018}, for which the output at initialization behaves as $1/\sqrt{h}$ where $h$ is the width. $h$ here is $128$ for the fully connected and the CNN architectures (\expModels).}
Such an inflation is not captured by the perceptron model. As a consequence, reasoning on the magnitude of the first gradient step may not be appropriate, and the finding (iii) above is not observed as shown in Fig. \figAlign.  
Yet, after inflation occurs, as for the perceptron the output still needs to grow even further in the good direction to overcome the noise from SGD, the more so the larger the training set. This effect is observed both for CNNs and fully-connected nets on different data sets, and in the latter case strongly correlates with performance \cite{dissectingSGD}. 

For fully connected networks learning MNIST, observations are indeed similar to the small margin case: (i,ii) a phase diagram suggesting three regimes affecting performance is shown in Fig. \figFCphase and (iv) a critical batch size $\Bc$ that again appears to follow a power law of the training set size, as $\Bc\sim P^{0.4}$ as shown in Fig. \figBcrit. Interestingly, using early-stopping or not does not affect performance, as shown in Fig. \figFCphaseES. It is consistent with our analysis of weight changes, which for the perceptron are similar at early stopping or at the end of training. 
For CNNs learning CIFAR10, the picture is different. Although three learning regimes may be identifiable in Fig. \figCNNphase, most of the dependence of performance with $\eta,B$ is gone when using early stopping as shown in Fig. \figCNNphaseES. It suggests that effects other than the growth of weights induced by SGD control performance in this example, as discussed below.

\section{Conclusion}
\label{sec:conclusions}

In deep networks, the effects of SGD on learning are known to depend on the size of the training set.
We used a simple toy model, the perceptron, to explain these observations and relate them to the difficulty of the task.  SGD noise increases the dependence of the predictor along incorrect input directions, a phenomenon that must be compensated by aligning more strongly toward the true function.  As a result, alignment and weight changes depend on both $T$ and $P$. If temperature is too small, this alignment is instead fixed by overcoming the margin and SGD is equivalent to GD. If the batch size is larger than a $P$-dependent $B^*$, the weight changes are instead governed by the first few steps of gradient descent. 
As one would expect, the alignment magnitude correlates with performance. It was observed in several cases in \cite{dissectingSGD}, and is also reflected by our observation that different alignment regimes correspond to different regimes of performance. In  \nnPlots,\ we discuss why it is so in the simple example of a teacher perceptron learnt by a multi-layer network. As shown in Fig. \figStripe, as weights align more strongly in the true direction as SGD noise increases, the network becomes less sensitive to irrelevant directions in input space, thus performing better. Yet, depending on the data structure, a strong alignment could be beneficial or not- a versatility of outcome that is observed  \cite{shallue2019, hoffer2017, dissectingSGD, andriushchenko2023}.
Obviously,  other effects of SGD independent on training set size may affect further performance, such as escaping saddles \cite{ge2015escaping, jordan2017saddles, daneshmand2018escaping, hu2019diffusionSGD, zhu2019anisotropic, benarous2022, florent2023unifyingSGD}, biasing the dynamics toward broader minima \cite{chaudhari2018, zhang2018energy, wu2018dynamical, xie2021diffusion, tu2021, tu2023,hinton1993keeping, hochreiter1997flat, baldassi2016unreasonable, keskar2016, neyshabur2017exploring, wu2017towards, entropysgd2019, baldassi2020shaping,andriushchenko2023} or finding sparser solutions  \cite{ganguli2023,flammarion2021,flammarion2023, Yu2023SGD}.  We have shown examples where the alignment effects appear to dominate, and others where SGD instead is akin to a regularization similar to early stopping, which is the case for the CNN in Fig. \figCNNphaseES -  a situation predicted in some theoretical approaches, see e.g. \cite{ganguli2023}.  Determining which effect of SGD most strongly affects performance given the structure of the task and the architecture is an important practical question for the future.
Our work illustrates that subtle aspects enter this equation - such as the size of the training set or the density of points near the decision boundary.

\matmethods{Experiments' code is available at \href{https://github.com/pcsl-epfl/regimes_of_SGD}{github.com/pcsl-epfl/regimes\_of\_SGD} \cite{sclocchiGitHub}.
}

\acknow{We thank F. Cagnetta, A. Favero, M. Geiger, B. Göransson, F. Krzakala, L. Petrini, U. Tomasini and L. Zdeborova for discussion. M.W. acknowledges support from the Simons Foundation Grant (No. 454953 Matthieu Wyart).}
\showmatmethods{} 
\showacknow{} 

\bibsplit[7]

\bibliography{bibliography}

\newpage
\onecolumn

\appendixstyle

\SItext
\hypertarget{app:expModels}{\section{Experimental models}}
\label{app:expModels}
\hypertarget{app:expModelsPercep}{\subsection{Perceptron model}}
\label{app:expModelsPercep}

\subsubsection{Loss and algorithm}
We train the perceptron with the hinge loss
\beqn
L(\w) &= \frac{1}{P} \sum_{\mu=1}^P (\ma-y^{\mu} f(\w,\x^{\mu}))^+,
\label{eq:percepLoss}
\eeqn
where $(x)^+ = \max(0,x)$, $\ma>0$ is the margin, $\{(\x^{\mu},y^\mu=y(\x^\mu))\}_{\mu=1,...,P}$ is the training set with size $P$. 
The gradient from the data point $(\x^{\mu},y^\mu)$ corresponds to
\beq
\nabla_\w\loss(\w,\x^\mu) = -\theta(\ma-y^{\mu} f(\w,\x^{\mu})) y^{\mu} \frac{\x^{\mu}}{\sqrt{d}},
\label{eq:grad}
\eeq
where $\theta(x)$ is the Heaviside step function, and the gradient of Eq. \ref{eq:percepLoss} is $\nabla_\w L(\w) = \frac{1}{P} \sum_{\mu=1}^P \nabla_\w \loss(\w,\x^\mu)$.
The fraction of training points giving a non-zero contribution to $L(\w)$ and $\nabla_\w L(\w)$ is given by
\beq
    \unf(\w) = \frac{1}{P} \sum_{\mu=1}^P \theta(\ma-y^{\mu} f(\w,\x^{\mu})).
    \label{eq:unfitted}
\eeq

The stochastic gradient descent update equation is:
\beq\label{eq:SGD}
    \w^{t+\dt} =\w^t + \frac{\dt}{B}\sum\limits_{\mu \in \Ba_t} \theta\lpa\ma-y^{\mu} f(\w,\x^{\mu})\rpa y^{\mu} \frac{\x^\mu}{\sqrt{d}}
\eeq
where $\Ba_t \subset \{1,...,P\}$ is the batch at time $t$ and $B$ is its size. 
The batch is randomly selected at each time step among all the $P$ data.
The learning rate $\dt$ is kept constant during training. The end of training is reached when $L(\w^{\tf})=\unf(\w^{\tf})=0$, with $\unf(\w) = \frac{1}{P} \sum_{\mu=1}^P \theta(\ma-y^{\mu} f(\w,\x^{\mu}))$.
Reported data are averaged over $5$ random initialization seeds.

In the limit $B\ll P$, the covariance of the mini-batch gradients is given by $\cov\lpa \w\rpa/B$ \cite{chaudhari2018}, with
\beqn
\begin{aligned}
    \cov\lpa \w\rpa &= \frac{1}{P}\sum_{\mu=1}^P\theta\lpa\ma-y^{\mu} f(\w,\x_{\mu})\rpa\ 
    \frac{\x^\mu \otimes \x^\mu}{d}\\
    &- \nabla_\w L(\w)\otimes \nabla_\w L(\w).
\end{aligned}
\label{eq:general_cov}
\eeqn

\subsection{Deep networks}
Given the width $h$, we initialize the weights with variance $1/\sqrt{h}$ for the hidden layers, and with variance $1/h$ in the output layer.\\
For the fully connected architecture, $h$ is the number of hidden neurons at each layer. We consider an architecture with $5$ hidden layers, gelu activation functions and $h=128$.\\
As deep convolutional network, we consider the MobileNet architecture \cite{mnasnet}. In this case, the number of channels, which changes along the layers, is proportional to $h$. We choose the width such that the first convolutional block has $128$ channels. The code in JAX/Haiku with the details of the implementation is available at \url{https://github.com/pcsl-epfl/regimes_of_SGD} \cite{sclocchiGitHub}.

Calling $\w^t$ the network parameters at time $t$ and $\x$ a datum, the output of the network is given by $f(\w^t,\x)$. We consider the prediction of the model on $\x$ to be $F(\w^t, \x) = f(\w^t,\x) - f(\w^0,\x)$, such that at initialization the predictor is unbiased. We train the network by minimizing the hinge-loss with margin $\ma$ as for the perceptron in Eq. \ref{eq:percepLoss}:
\beqn
L(\w^t) &= \frac{1}{P} \sum_{\mu=1}^P (\ma-y^{\mu} F(\w^t,\x^{\mu}))^+.
\label{eq:netLoss}
\eeqn

We train with SGD at constant $\dt$ and $B$ until reaching $L(\w^{\tf})=\unf(\w^{\tf})=0$.
Reported data are averaged over $5$ random initialization seeds.

\hypertarget{app:extensions}{\section{Extensions to momentum and weight decay}}
\label{app:extensions}

In practice, deep learning models are often trained using the SGD algorithm with the addition of momentum and/or weight decay \cite{sutskever2013}.
\subsection{Momentum}
In the case of momentum, the step update is modified as 
\begin{align}
\begin{aligned}
    \vel^{t+\eta} &= m \vel^{t} - \frac{\dt}{B} \sum_{\mu\in\Ba_t} \nabla_\w l(\w^t, \x^\mu),\\
    \w^{t+\dt} &= \w^t + \vel^{t+\eta},
\end{aligned}
\end{align}
where $m\in [0,1]$ is the momentum coefficient and $m=0$ corresponds to standard SGD. Previous work \cite{mandt2017stochastic, shallue2019} observed that momentum induces an effective learning rate $\dt_{\rm eff} = \frac{\dt}{1-m}$ and thus an effective SGD temperature $T_{\rm eff}=\dt_{\rm eff}/B$.
Fig. \ref{fig:mom_wd}-(a) shows the dynamics of the perceptron weights $w_1$ and $\|\w_\perp\|$ when trained with SGD with momentum ($m=0.9$), in a setting identical to Fig. 2 in the main text. Although the presence of momentum modifies the weight trajectories with respect to Fig. 2, we still observe a small-batch regime where the final point is determined by the effective SGD temperature $T_{\rm eff}=\dt_{\rm eff}/B$, and a large-batch regime dominated by the first training steps and controlled by the effective learning rate $\dt_{\rm eff}$. This observation suggests that the addition of momentum does not change the structure of the phase diagram up to a rescaling of the learning rate.

\hypertarget{app:extensionsWD}{\subsection{Weight decay}}
\label{app:extensionsWD}
In the case of weight decay, the SGD update is modified as
\begin{align}
    \w^{t+\dt} &= \w^t - \frac{\dt}{B} \sum_{\mu\in\Ba_t} \nabla_\w l(\w^t, \x^\mu) - \dt \Lambda \w^t,
    \label{eq:wd}
\end{align}
where $\Lambda$ is the weight decay coefficient. For $\Lambda>0$, the dynamics never stops and the training loss does not reach a constant zero value. Therefore, at long times the perceptron weights fluctuate around a stationary minimum.
The coefficient $\Lambda$ defines a time scale 
\beq
t_\Lambda = \Lambda^{-1}.
\eeq 
In the following, we consider small values of $\Lambda$ such that $t_\Lambda$ is comparable with the training time needed to reach zero loss in the absence of weight decay.

\subsubsection{Small batch}
\begin{figure}[H]
    \centering
    \begin{tikzpicture}
        \node[anchor=north west,inner sep=0pt] at (0,0){    \resizebox{.47\textwidth}{!}{
        \input{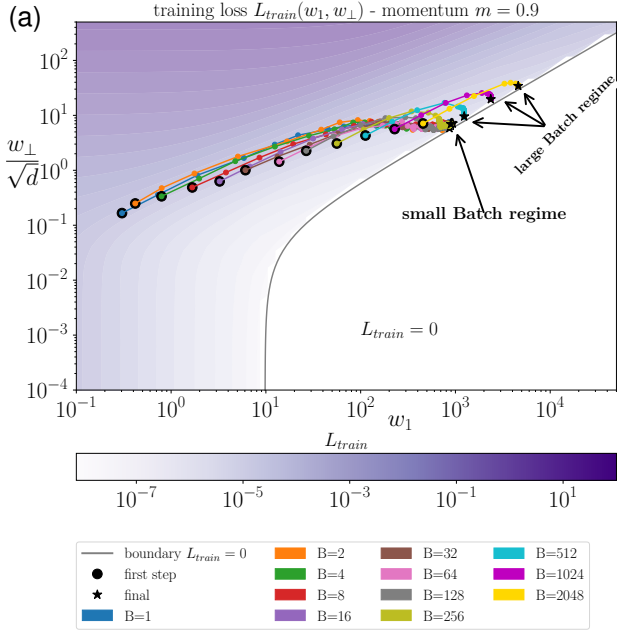}}};
        \node[font=\sffamily\Large] at (3ex,-2.5ex) {(a)};
    \end{tikzpicture}
    \hspace{.5cm}
    \begin{tikzpicture}
        \node[anchor=north west,inner sep=0pt] (fig) at (0,0){\resizebox{.48\textwidth}{!}{
        \input{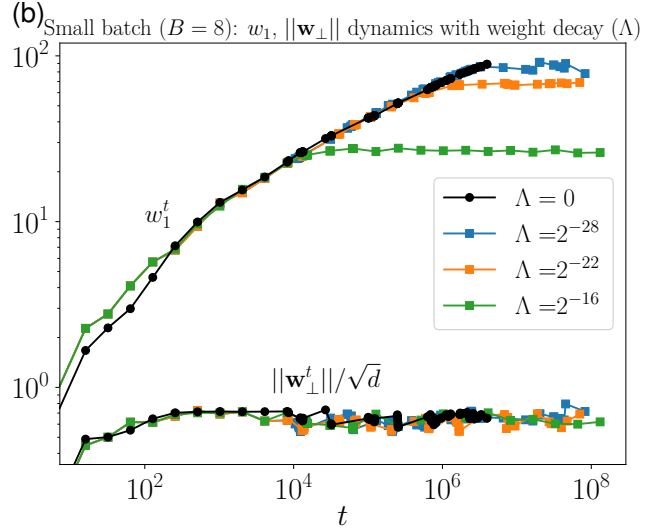}}};
        \node[font=\sffamily\Large] at (3ex,-0.5ex) {(b)};
        \node[] at (3ex,-58ex) {}; 
    \end{tikzpicture}
    \caption{
    \textbf{(a) Dynamical trajectories of SGD with momentum \cite{sutskever2013} (momentum coefficient $m=0.9$) in the $(w_1, w_\perp)$ at fixed $\dt/B=2$ and varying batch size as in caption.} The setting is identical to Fig. 2 in the main text. For small batch ($B<512$), the final point of the dynamics is determined by the effective temperature $T_{eff}=\frac{\dt}{(1-m) B}$. In fact, with respect to Fig. 2, the final point of $w_\perp\propto T_{eff}$ is shifted upwards of a factor $1/(1-m)=10$. For large batch ($B \gtrsim 512$), the dynamics is dominated by the first steps.
    \textbf{(b) Dynamics of the perceptron weights using SGD with weight decay, varying $\Lambda$, fixed $B=8$, $\dt=16$, $\chi=1$, $d=128$, $P=16384$, $\ma=2^{-7}$.} $\|\w_\perp\|$ is unaffected by $\Lambda$ and converges to the same value proportional to $T=\dt/B$. The growth of $w_1$, instead, is interrupted at a time scale that increases for decreasing $\Lambda$.
    }
    \label{fig:mom_wd}
\end{figure}

Fig. \ref{fig:mom_wd}-(b) shows that weight decay affects the dynamics of $w_1^t$ at the end of training, while it does not affect $\|\w_\perp^t\|$, which reaches the same stationary value as in the $\Lambda=0$ case. Therefore, in the noise-dominated regime, at long times $\|\w_\perp\|\propto T\sqrt{d}$ (Eq. 8 in the main text) as in the absence of weight decay.
Since the noise-dominated phase is determined by the magnitude of $\|\w_\perp\|$ with respect to the margin $\ma$ (Sec. 1.B in the main text), then the presence of weight decay does not change the critical line $T_c \propto \ma$ (Eq. 13 in the main text) in the phase diagram, separating the GD and the noise-dominated-SGD phases.
The effect of $\Lambda$ on the stationary value of $w_1$ is obtained by adding the weight decay term to Eq. 7 in the main text for $dw_1^t$. In the noise-dominated regime, we have
\beq
    dw_1^t = dt \lpa\frac{\|\w_\perp^t\|}{w_1^t}\rpa^{\chi+2} \frac{c_1}{\sqrt{d}} - dt\ \Lambda w_1^t.
    \label{eq:w1_wd}
\eeq
Using $\|\w_\perp^t\|\propto T$, the right hand side of Eq. \ref{eq:w1_wd} vanishes for $w_1^t$ reaching the stationary value $w_{1,\Lambda}$:
\beq
    w_{1,\Lambda} \propto T^{\frac{\chi+2}{\chi+3}} \Lambda^{-\frac{1}{\chi+3}},
    \label{eq:w1_T_L}
\eeq
which is consistent with the empirical data in the inset of Fig. \ref{fig:wd_test}-(a). 
The scaling of Eq. \ref{eq:w1_T_L} cannot be valid up to $\Lambda\rightarrow 0$, since we know that, for $\Lambda=0$, the final value of $w_1$ is given by $w_{1,\Lambda=0} \sim TP^\ec$ (Eq. 11 in the main text) with  $\ec=\frac{1}{\chi+1}$. Therefore, there is a cross-over between the two behaviors when $w_{1,\Lambda}\sim w_{1,\Lambda=0}$, that is for $\Lambda\sim\Lambda^*$ with
\beq
    \Lambda^* \sim T^{-1} P^{-b},
    \label{eq:L_star}
\eeq
where $b=1+\frac{2}{1+\chi}$. For $\Lambda\gg \Lambda^*$, the final minimum of the dynamics is determined by $\Lambda$, while for $\Lambda\ll \Lambda^*$ it is determined by $P$.
Equivalently, the condition of Eq. \ref{eq:L_star} can be interpreted in terms of time scales. In fact, the training time in absence of weight decay is given by $\tcross\sim T P^b$ (Eq. 11 in the main text) and the condition of Eq. \ref{eq:L_star} is equivalent to $t_{\Lambda^*}\sim \tcross$.\\
For $\Lambda\gg \Lambda^*$, the test error $\epsilon$ increases with $\Lambda$ as
\beq
    \epsilon \sim \int_{0}^{\frac{\|\w_\perp\|}{w_1}} x_1^{\chi} dx_1 \sim \lpa\frac{\|\w_\perp\|}{w_1}\rpa^{\chi+1} \sim \lpa T\Lambda \rpa^{\frac{\chi+1}{\chi+3}},
    \label{eq:err_L}
\eeq
where we have used $\|\w_\perp\|\propto T$ and Eq. \ref{eq:w1_T_L}. This prediction is consistent with the data in the main panel of Fig. \ref{fig:wd_test}-(a), where we observe that, in this model, $\Lambda>0$ has no benefit on the test error, since no label noise is present.

\subsubsection{Large batch}
\begin{figure}[H]
    \centering
    \begin{tikzpicture}
        \node[anchor=north west,inner sep=0pt] at (0,0){    \resizebox{.46\textwidth}{!}{
        \input{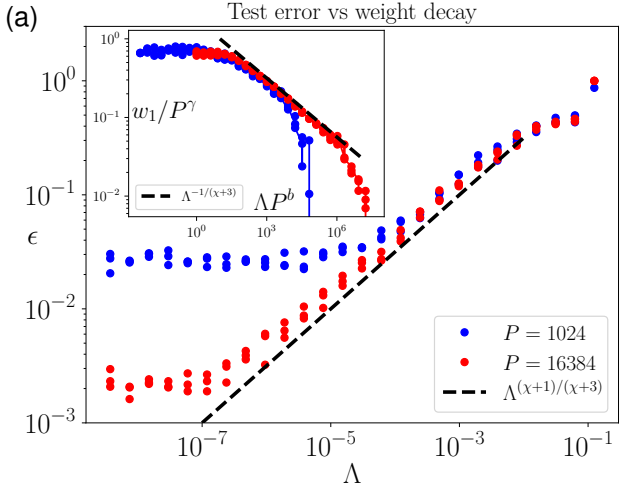}}};
        \node[font=\sffamily\Large] at (1ex,-2.5ex) {(a)};
    \end{tikzpicture}
    \hspace{.5cm}
    \begin{tikzpicture}
        \node[anchor=north west,inner sep=0pt] (fig) at (0,0){\resizebox{.46\textwidth}{!}{
        \input{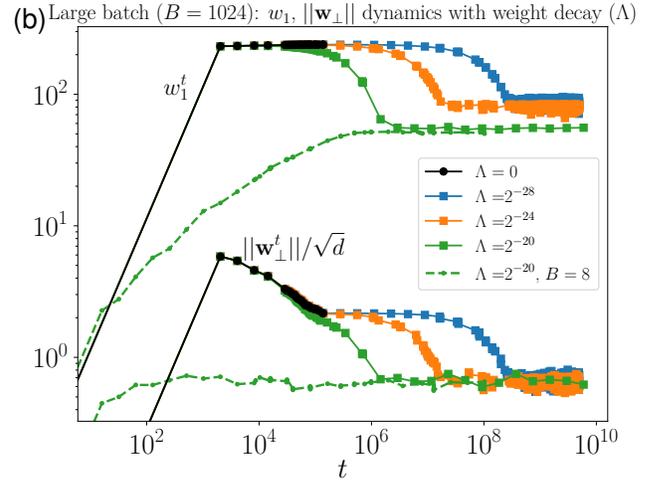}}};
        \node[font=\sffamily\Large] at (1ex,-2.5ex) {(b)};
    \end{tikzpicture}
    \caption{
    \textbf{(a) In the perceptron model, the test error increases with increasing weight decay $\Lambda$ (main panel) when $\Lambda>\Lambda^*$, where $\Lambda^*$ depends on the training set size $P$. This is due to the reduction of the final value of the informative weight $w_1$ (inset), which exhibits a cross-over from $w_1\sim P^\ec$, $\ec=1/(1+\chi)$, for $\Lambda\ll\Lambda^*$ to $w_1\sim \Lambda^{-1/(\chi+3)}$ for $\Lambda\gg\Lambda^*$, with $\Lambda^*\sim P^b$, $b=1+2/(1+\chi)$.} In the considered model, weight decay does not benefit performance since there is no label noise in the training samples. Data obtained in the noise-dominated regime (same parameters as Fig. \ref{fig:mom_wd}-(b)).
    \textbf{(b) For large batch size ($B=1024$) and $\Lambda>0$, the first-step-dominated regime disappears at a time-scale proportional to $\Lambda^{-1}$.} For small $\Lambda$, the first-step dynamics of $w_1$ is the same as in the absence of weight decay ($\Lambda=0$). At a time scale proportional to $\Lambda^{-1}$, instead, $w_1$ converges to a steady state depending on $\Lambda$. This is the same value as the one obtained for small batch and same SGD temperature $T=\dt/B$ (green dashed line).
    The dynamics of $\|\w_\perp\|$ has the same phenomenology as that of $w_1$, but its steady state only depends on the SGD temperature $T$. Data with fixed $\dt/B=2$ ($B=1024$ for the full lines, $B=8$ for the dashed green lines), other parameters as in Fig. \ref{fig:mom_wd}-(b).
    }
    \label{fig:wd_test}
\end{figure}

Fig. \ref{fig:wd_test}-(b) shows that, for small value of $\Lambda$, the first-step dynamics of $w_1^t$ and $\|\w_\perp^t\|$ is the same as in the case $\Lambda=0$. Instead, at a time-scale $t\sim t_\Lambda$, both $w_1^t$ and $\|\w_\perp^t\|$ converge to a stationary minimum, where only the final value of $w_1$ is affected by $\Lambda$. 
Fig. \ref{fig:wd_test}-(b) shows that this solution is the same as the small batch case, which depends on $\Lambda$ and the SGD temperature $T=\dt/B$. 
Therefore, in the presence of weight decay, the first-step-dominated phase of SGD disappears in favor of the noise-dominated one. However, it can be recovered by performing early-stopping at training times shorter than $t_\Lambda$.

\hypertarget{app:dnnExp}{\section{Deep networks experiments}}
\label{app:dnnExp}

\subsection{Escaping the lazy regime when the margin $\ma$ is small}
\begin{figure}[H]
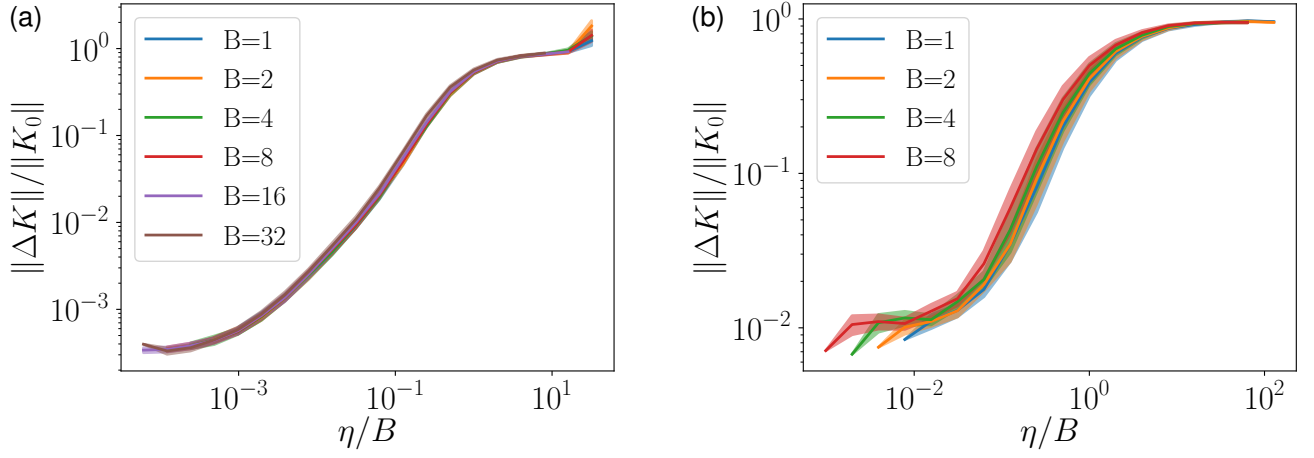

    \centering
    \begin{tikzpicture}
        \node[anchor=north west,inner sep=0pt] at (0,0){    \resizebox{.47\textwidth}{!}{
        \input{figures/FC_5L-mnist-Tplot_kernel_change_sqrt-alpha32768.0_P32768.pgf}}};
        \node[font=\sffamily\Large] at (3ex,-2.5ex) {(a)};
    \end{tikzpicture}
    \hspace{.5cm}
    \begin{tikzpicture}
        \node[anchor=north west,inner sep=0pt] at (0,0){    \resizebox{.47\textwidth}{!}{
        \input{figures/MNAS-cifar-Tplot_kernel_change_sqrt-alpha32768.0_P16384.pgf}}};
        \node[font=\sffamily\Large] at (3ex,-2.5ex) {(b)};
    \end{tikzpicture}
    \caption{
    \textbf{Neural tangent kernel (NTK) variation at the end of training relative to its initialization magnitude $\|\Delta K\|/\|K_0\|$: increasing SGD temperature $T=\dt/B$ induces a larger variation of the NTK, escaping the lazy training regime.}
    (a) Fully-connected (5-hidden layers) on parity MNIST, $\ma=2^{-15}$, $P=32768$.
    (b) CNN on CIFAR10, $\ma=2^{-15}$, $P=16384$.
    }
    \label{fig:ntk}
\end{figure}

For very small $\ma$ and training with gradient flow, networks fit the data with tiny changes of their weights. In this lazy training regime, neural networks are equivalent to kernel methods with the neural tangent kernel (NTK) $K(\x^\mu,\x^\nu) = \nabla_\w f(\w,\x^\mu)\cdot \nabla_\w f(\w,\x^\nu)$ frozen at initialization \cite{jacot2018, chizat2019lazy, geiger2020disentangling}. This is the case if the magnitude of the variation of the NTK at the end of training, $\|\Delta K\| = \| K_{\tf}-K_0 \|$ is much smaller than its initialization magnitude $\|K_0\|$.
Fig. \ref{fig:ntk} shows that, even with tiny $\ma$, the relative NTK variation $\|\Delta K\|/\|K_0\|$ increases with increasing SGD temperature $T=\dt/B$, therefore escaping the lazy training regime. We estimate the NTK magnitude through the Frobenius norm of its Gram matrix on $128$ training points.\\

\subsection{Improving generalization by adapting to the structure of the task}
\begin{figure}[H]
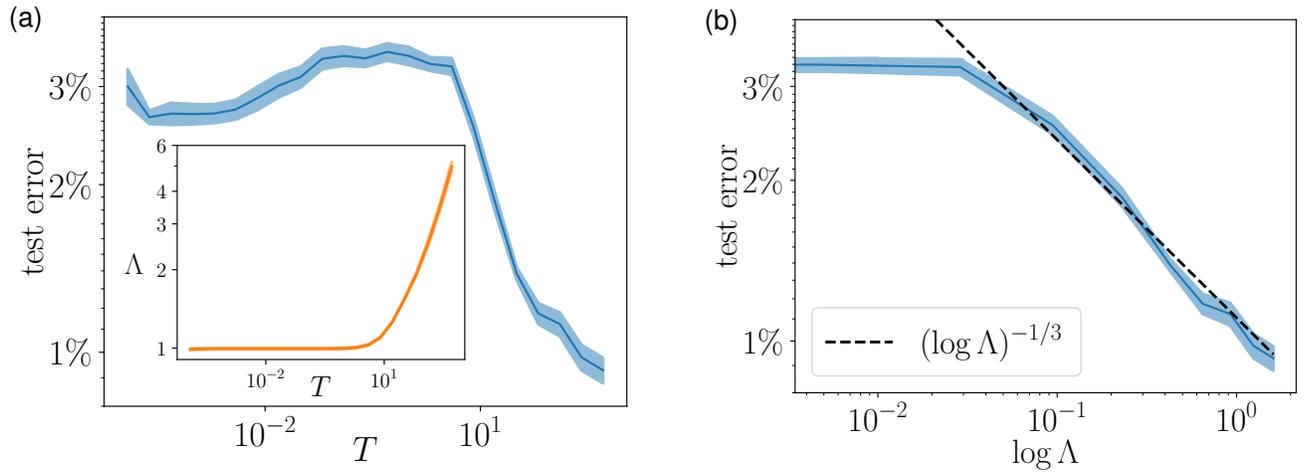

    \centering
    \begin{tikzpicture}
        \node[anchor=north west,inner sep=0pt] at (0,0){    \resizebox{.48\textwidth}{!}{
        \input{figures/FC_5L-depleted_sign_chi1_d128-Tplot_B1_test_err-alpha32768_P8192.pgf}}};
        \node[font=\sffamily\Large] at (3ex,-2.5ex) {(a)};
    \end{tikzpicture}
    \hspace{.5cm}
    \begin{tikzpicture}
        \node[anchor=north west,inner sep=0pt] at (0,0){    \resizebox{.46\textwidth}{!}{
        \input{figures/FC_5L-depleted_sign_chi1_d128-test_err_vs_logampli-alpha32768_P8192.pgf}}};
        \node[font=\sffamily\Large] at (3ex,-2.5ex) {(b)};
    \end{tikzpicture}
    \caption{
    \textbf{Increasing the SGD temperature, deep networks adapt to the structure in the data.}
    Fully-connected (5-hidden layers) architecture trained on linearly separable data: labels $y=\sign(x_1)$, data distribution Eq. 1 in the main text with $\chi=1$, dimension $d=128$, hinge loss margin $\ma=2^{-15}$, $P=8192$.
    (a) Main: test error at the end of training. Increasing the SGD temperature $T$, the test error starts decreasing when $T$ is larger than some threshold. Inset: this threshold corresponds to the network adapting to the anisotropy of the task, corresponding to $\Lambda>1$ (Eq. \ref{eq:amplifactor}).
    (b) The test error decreases logarithmically with increasing $\Lambda>1$.
    }
    \label{fig:stripe}
\end{figure}

Escaping the kernel regime, the network can adapt to the structure of the data/task and improve generalization. To observe this fact, we train a fully-connected (5-hidden layers) architecture on the linearly separable task defined for the perceptron in section 1.A of the main text, whose class labels are given by $y=\sign(x_1)$. With random initialization, the network has no preferred direction in input space and the corresponding NTK is isotropic. Since the task only depends on $x_1$, we measure how much the network is sensitive to this direction with the amplification factor $\Lambda$:
\beq
    \Lambda^2 = (d-1)\frac{\sum_n w_{n,1}^2}{\sum_n w_{n,\perp}^2},
    \label{eq:amplifactor}
\eeq
where the sum $\sum_n$ is over the first layer neurons. At initialization, $\Lambda=1$ corresponds to an isotropic NTK. If the network escapes lazy training, then $\Lambda>1$ at the end of training, signaling a growth of the weights in the informative input direction $x_1$. Fig. \ref{fig:stripe} shows that, even with a small hinge loss margin $\ma$, increasing SGD temperature leads to a regime with $\Lambda>1$, and correspondingly the test error starts improving. Becoming more sensitive to the informative direction, the network is reducing the dimensionality of the task.

\subsection{Experiments with large margin $\ma$}
\begin{figure}[H]
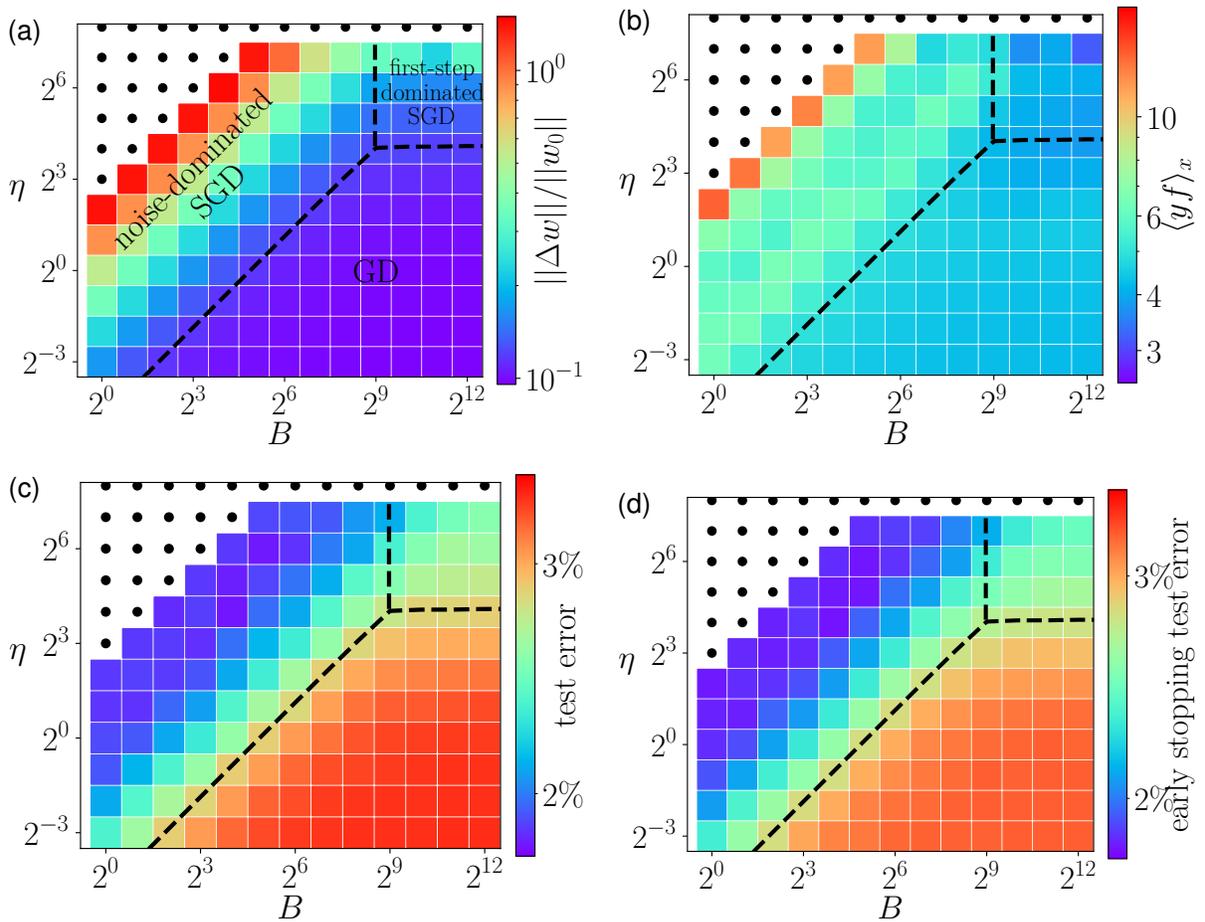

    \centering
    \begin{tikzpicture}
        \node[anchor=north west,inner sep=0pt] at (0,0){    \resizebox{.45\textwidth}{!}{
        \input{figures/FC_5L-mnist-phase-dt_B_rel_dw-alpha1.0_P16384.pgf}}};
        \node[font=\sffamily\Large] at (3ex,-2.5ex) {(a)};
    \end{tikzpicture}
    \begin{tikzpicture}
        \node[anchor=north west,inner sep=0pt] at (0,0){    \resizebox{.45\textwidth}{!}{
        \input{figures/FC_5L-mnist-phase-dt_B_yf-alpha1.0_P16384.pgf}}};
        \node[font=\sffamily\Large] at (3ex,-2.5ex) {(b)};
    \end{tikzpicture}
    \begin{tikzpicture}
        \node[anchor=north west,inner sep=0pt] at (0,0){    \resizebox{.45\textwidth}{!}{
        \input{figures/FC_5L-mnist-phase-dt_B_test_err-alpha1.0_P16384.pgf}}};
        \node[font=\sffamily\Large] at (3ex,-2.5ex) {(c)};
    \end{tikzpicture}
    \begin{tikzpicture}
        \node[anchor=north west,inner sep=0pt] at (0,0){    \resizebox{.45\textwidth}{!}{
        \input{figures/FC_5L-mnist-phase-dt_B_err_ES-alpha1.0_P16384.pgf}}};
        \node[font=\sffamily\Large] at (3ex,-2.5ex) {(d)};
    \end{tikzpicture}

    \caption{\textbf{SGD phase diagrams in batch size ($B$) and learning rate ($\dt$) for fully-connected (5-hidden layers) on parity MNIST, $\ma=1$, $P=16384$.}
    The black dots represent diverging trainings due to too large $\dt$.
    \textbf{(a) The weight variation at the end of training, relative to their initialization, distinguishes the different dynamical regimes.} The diagonal dashed line for small $B$ and the horizontal one for large $B$ are given by $\frac{\|\Delta \w\|}{\|\w_0\|}$ being a factor $1.3$ larger of its value for $\dt\rightarrow 0$. 
    The diagonal and the horizontal dashed lines meet at the critical batch size $\Bc$ (vertical dashed line).
    \textbf{(b) The alignment $\langle y(\x)f(\x) \rangle_\x$ at the end of training.} 
    \textbf{(c) The test error at the end of training is best in the noise-dominated regime.} In the first-step-dominated regime, performance still improves with respect to the GD case, but not as much as in the noise-dominated one.
    \textbf{(d) Using early stopping, the dependence of the test error from $\dt$ and $B$ is the same as at the end of training (panel (c)).} Although this procedure is slightly beneficial for GD, the performance still depends on $\dt$ and $B$ as at the end of training (panel (c)), with the optimal performance achieved for noise-dominated SGD. This implies that its benefit is not just avoiding over-fitting, differently from the CNN case (Fig. \ref{fig:CNNfeature}).
    }
    \label{fig:FCfeature}
\end{figure}

\begin{figure}[H]
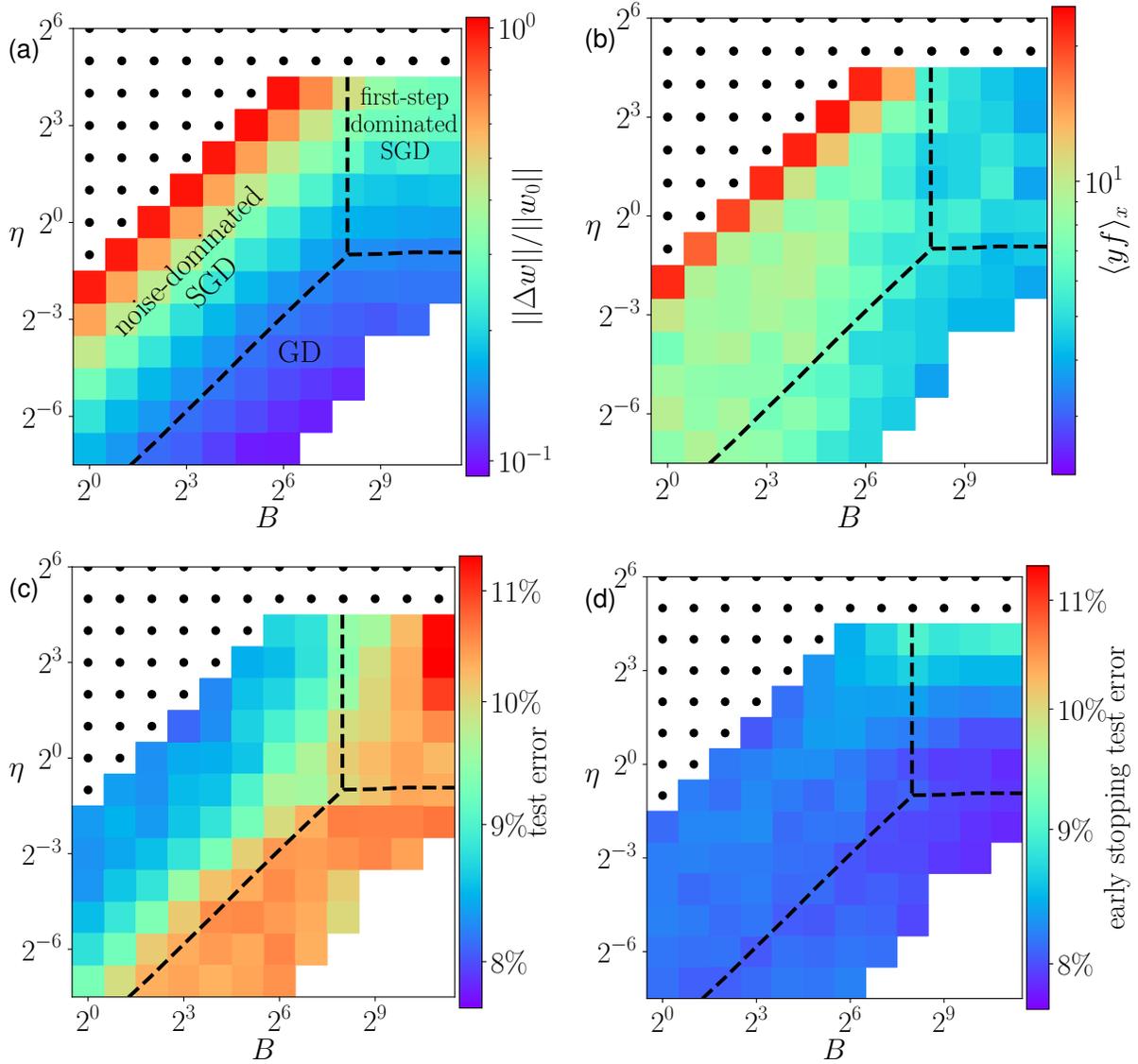

    \centering
    \begin{tikzpicture}
        \node[anchor=north west,inner sep=0pt] at (0,0){    \resizebox{.45\textwidth}{!}{
        \input{figures/MNAS-cifar-phase-dt_B_rel_dw-alpha1.0_P16384.pgf}}};
        \node[font=\sffamily\Large] at (3ex,-4.5ex) {(a)};
    \end{tikzpicture}
    \begin{tikzpicture}
        \node[anchor=north west,inner sep=0pt] at (0,0){    \resizebox{.45\textwidth}{!}{
        \input{figures/MNAS-cifar-phase-dt_B_yf-alpha1.0_P16384.pgf}}};
        \node[font=\sffamily\Large] at (3ex,-4.5ex) {(b)};
    \end{tikzpicture}
    \begin{tikzpicture}
        \node[anchor=north west,inner sep=0pt] at (0,0){    \resizebox{.45\textwidth}{!}{
        \input{figures/MNAS-cifar-phase-dt_B_test_err-alpha1.0_P16384.pgf}}};
        \node[font=\sffamily\Large] at (3ex,-4.5ex) {(c)};
    \end{tikzpicture}
    \begin{tikzpicture}
        \node[anchor=north west,inner sep=0pt] at (0,0){    \resizebox{.45\textwidth}{!}{
        \input{figures/MNAS-cifar-phase-dt_B_err_ES-alpha1.0_P16384.pgf}}};
        \node[font=\sffamily\Large] at (3ex,-4.5ex) {(d)};
    \end{tikzpicture}

    \caption{\textbf{SGD phase diagrams in batch size ($B$) and learning rate ($\dt$) for CNN on CIFAR10, $\ma=1$, $P=16384$.} 
    The black dots represent diverging trainings due to too large $\dt$.
    \textbf{(a) The weight variation at the end of training, relative to their initialization, distinguishes the different dynamical regimes.} The diagonal dashed line for small $B$ and the horizontal one for large $B$ are given by $\frac{\|\Delta \w\|}{\|\w_0\|}$ being a factor $1.5$ of its value for $\dt\rightarrow 0$. 
    These two lines meet at the critical batch size $\Bc$ (vertical dashed line).
    \textbf{(b) Alignment $\langle y(\x)f(\x) \rangle_\x$ at the end of training.} 
    \textbf{(c) The test error at the end of training is best in the noise-dominated regime.} In the first-step-dominated regime, performance is comparable to the GD regime and degrades for the largest batch size.
    \textbf{(d) Using early stopping, the test error for noise-dominated SGD and GD is the same.} We observe that with this procedure GD achieves as good performance as noise-dominated SGD, implying that its benefit (panel (c)) in this setting corresponds to an effective regularization against over-fitting (differently from the fully-connected architecture in Fig. \ref{fig:FCfeature}).
    In the first-step dominated SGD, performance is slightly degraded with respect to GD, but it is still better than the end of training (panel (c)).
    }
    \label{fig:CNNfeature}
\end{figure}

Figs. \ref{fig:FCfeature} and \ref{fig:CNNfeature} show the phase diagrams for (a) weight variation of the end of training relative to their initialization; (b) alignment at the end of training; (c) test error at the end of training; (d) test error with early stopping. Measuring the test error along the training dynamics, we define the early-stopping point as the one corresponding to the best test error.

\begin{figure}[H]
    \centering
    \begin{tikzpicture}
        \node[anchor=north west,inner sep=0pt] at (0,0){    \resizebox{.5\textwidth}{!}{
        \input{figures/FC_mnist-batch-alignment_dyn-alpha1.0-eta64-P16384.pgf}}};
        \node[font=\sffamily\Large] at (10ex,-4.5ex) {(a)};
    \end{tikzpicture}
    \begin{tikzpicture}
        \node[anchor=north west,inner sep=0pt] at (0,0){    \resizebox{.45\textwidth}{!}{
        \input{figures/FC_mnist-batch-rel_dw-alpha1.0.pgf}}};
        \node[font=\sffamily\Large] at (10ex,-4.5ex) {(b)};
    \end{tikzpicture}
    \caption{\textbf{Fully-connected (5-hidden layers) on parity MNIST, $\ma=1$, $P=2^{14}=16384$.}\\
    \textbf{(a) Dynamics of the alignment during training, fixed $\dt=64$.} For $\ma=1$, the dynamics of the alignment is different from the small $\ma$ case (Fig. 3 in main text).
    This behavior is different from the perceptron case, where changing $\ma$ does not change the dynamics, up to a rescaling of the learning rate.
    \textbf{(b) Relative weight variation at the end of training vs batch size}.
    Inset: $\Delta w$ depends on $\dt$ and $B$ for small $B$, while it only depends on $\dt$ for large $B$.
    Main: rescaling $B$ by $P^{0.4}$ aligns the cross-over between small and large batch at different $P$s, suggesting a critical batch size $\Bc\propto P^{0.4}$.
    This is a different exponent with respect to the small $\ma$ case, where $\Bc \propto P^{0.2}$ (Fig. 4-(b) in main text).
    }
    \label{fig:FC_dynamics_feature}
\end{figure}

\begin{figure}[H]
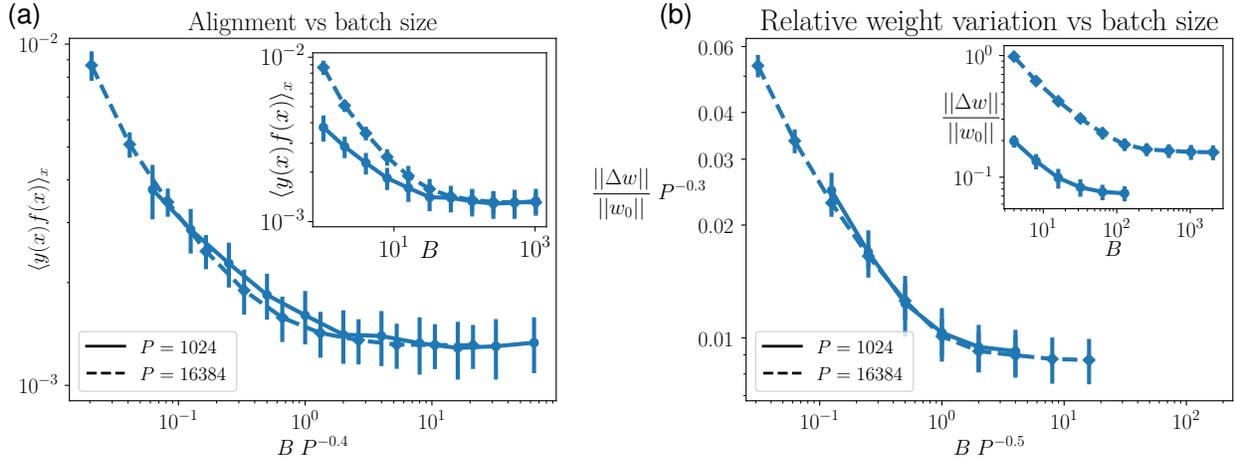

    \centering
    \begin{tikzpicture}
        \node[anchor=north west,inner sep=0pt] at (0,0){    \resizebox{.42\textwidth}{!}{
        \input{figures/MNAS_cifar-batch-yf-alpha32768.0.pgf}}};
        \node[font=\sffamily\Large] at (2ex,-1.5ex) {(a)};
    \end{tikzpicture}
    \begin{tikzpicture}
        \node[anchor=north west,inner sep=0pt] at (0,0){    \resizebox{.5\textwidth}{!}{
        \input{figures/MNAS_cifar-batch-rel_dw-alpha1.0.pgf}}};
        \node[font=\sffamily\Large] at (10ex,-1.5ex) {(b)};
    \end{tikzpicture}

    \caption{\textbf{CNN on CIFAR10: the critical batch size depends on $P$.} \textbf{(a) Alignment at the end of training vs batch size, $\ma=2^{-15}$, $\dt=32$.} Inset: The alignment depends on $B$ only for small $B$, and the critical batch size depends on the size of the training set $P$.
    Main: rescaling $B$ by $P^{0.4}$ aligns the cross-over between small and large batch at different $P$s, suggesting a critical batch size $\Bc\propto P^{0.4}$.
    \textbf{(b) Relative weight variation at the end of training vs batch size, $\ma=1$, $\dt=1$.}
    Inset: $\Delta w$ depends on $B$ only for small $B$.
    Main: rescaling $B$ by $P^{0.5}$ aligns the cross-over between small and large batch at different $P$s, suggesting a critical batch size $\Bc\propto P^{0.5}$.
    }
    \label{fig:CNN_batch}
\end{figure}

\hypertarget{app:sdeSol}{\section{Summary of the solution of the online stochastic differential equation}}
\label{app:sdeSol}
The equation for $w_1^t$ is obtained by projecting the online SDE of $\w^t$ on $\ewt$. To obtain the equation for $\|\w_\perp^t\|$, we need to apply Ito's lemma to $d\w^t_\perp$ (Sec. \ref{app:sde}), which adds a deterministic contribution 
to $d\|\w_\perp^t\|$.

We obtain
\beqn
\begin{aligned}
    dw_1^t &= dt\ g_1^t + \sqrt{\frac{T}{d}} \sigma_1^t d\tilde{W}_1^t\\
    d\|\w_\perp^t\| &= dt\ \lsq g_\perp^t + \frac{T}{2\|\w_\perp^t\|} 
    \unf^t \rsq
    + \sqrt{\frac{T}{d}} \sigma_2^t d\tilde{W}_2^t
\end{aligned}
\label{eq:dynamics_app}
\eeqn
where the expressions for $g_1^t$, $g_\perp^t$, $\unf^t$, 
$\sigma_{1}^t$, $\sigma_{2}^t$, are functions of the time-dependent ratios
\beq
\ampli=\frac{w_1^t}{\|\w_\perp^t\|},\quad
\aw = \frac{\ma\sqrt{d}}{\|\w_\perp^t\|},
\eeq
and are reported in Sec. \ref{app:sde}.\\

\paragraph{Deterministic terms}
Inspecting the expressions of $g_1$, $g_\perp$ and $\unf^t$ (Sec. \ref{app:sde}) that appear in the deterministic part of Eq. \ref{eq:dynamics_app}, we make the following observations:
\begin{itemize}
    \item $g_1^t$ is always positive and makes the informative weight $w_1$ grow in time.
    \item $g_\perp^t$ is a negative term that pulls $\|\w_\perp^t\|$ to $0$. Conversely, a deterministic force coming from SGD noise and proportional to the fraction of unfitted points $\unf^t$ tends to make $\|\w_\perp^t\|$ larger.
\end{itemize}
From the balancing of $g_\perp^t$ and $\frac{T}{\|\w_\perp^t\|} \unf^t$, we expect the growth of $\|\w_\perp^t\|$ to be bounded. $w_1^t$, instead, grows indefinitely since $g_1^t$ only vanishes for $\ampli\rightarrow \infty$, which defines the fixed point of the dynamics.

\paragraph{Stochastic fluctuations}
The stochastic part of Eq. \ref{eq:dynamics_app} gives fluctuations around the solutions of the deterministic part that become negligible for $d\gg 1$. To show this point, in the following we consider solutions of Eq. \ref{eq:dynamics_app} of the form
\beq
w_1^t = \hat{w}_1^t + z_1^t, \quad
\|\w_\perp^t\| = \|\hat{\w}_\perp^t\| + z_\perp^t
\label{eq:ansatz}
\eeq
where $\hat{w}_1^t$, $\|\hat{\w}_\perp^t\|$ are solutions to the deterministic terms of Eq. \ref{eq:dynamics_app} and $z_1^t$, $z_\perp^t$ are the fluctuating parts due to the stochasticity in the dynamics. By expanding Eq. \ref{eq:dynamics_app} with respect to $\frac{z_1^t}{w_1^t}$ and $\frac{z_\perp^t}{\|\hat{\w}_\perp^t\|}$, we obtain a time-dependent Ornstein–Uhlenbeck process:

\beq
d \mathbf{z}^t = dt\ \mathbf{A}^t\ \mathbf{z}^t + \bm{\sigma}^t\ d\tilde{\mathbf{W}}^t,
\label{eq:noise_dyn}
\eeq
with the vector 
\beq
\mathbf{z}^t = 
\lpa\begin{matrix}
z_1^t\cr
z_\perp^t
\end{matrix}\rpa
.
\eeq
$\mathbf{A}^t$ and $\bm{\sigma}^t$ are matrices that depend on time through $\hat{w}_1^t$ and $\|\hat{\w}_\perp^t\|$. 

In Sec. \ref{app:stoch_terms} we study this process in the asymptotic regime $\ampli \rightarrow \infty$. We obtain that
\beq
    \langle {z_\perp^t}^2 \rangle^\frac{1}{2} / \hat{w}_\perp^t = \O\lpa d^{-1/2}\rpa, \quad
    \langle {z_1^t}^2 \rangle^\frac{1}{2} / \hat{w}_1^t = \O\lpa d^{-1/2}\rpa,
\eeq
which are small for $d\gg 1$.
This shows that the deterministic solution $\hat{w}_1^t$, $\|\hat{\w}_\perp^t\|$ is stable to noise fluctuations when $d\gg 1$.

\hypertarget{app:asy_sol}{\subsection{High temperature asymptotic solution}}
\label{app:asy_sol}
The asymptotic solution for large times 
in the high-temperature regime corresponds to the limits $\ampli\gg 1$ and $\aw\ll 1$ (Sec. 1.B in the main text).
In this limit, following the expansions of Eqs. \ref{eq:g1_asymp}, \ref{eq:gp_asymp}, \ref{eq:n_asymp-app}, the gradient terms become
\beq
    g_1^t = \frac{c_1}{\sqrt{d}} \ampli^{-\chi-2}, \quad
    g_\perp^t = -\frac{\ampli^{-\chi-1}}{\sqrt{2\pi d}}, \quad
    \unf^t = c_\unf \ampli^{-\chi-1},
    \label{eq:grad_asympt}
\eeq
with $c_1 = \frac{1+\chi}{\sqrt{2\pi}(2+\chi)} + \O\lpa \aw\rpa$ and $c_\unf = \frac{\Gamma(1+\frac{\chi}{2})}{2\sqrt{\pi}\Gamma(\frac{3+\chi}{2})} + \O\lpa \aw \rpa$.
The equations of motion for the deterministic part become:
\beqn
\begin{aligned}
    d\hat{w}_1^t &= dt \frac{c_1}{\sqrt{d}} \ampli^{-\chi-2}, \\
    d\|\hat{\w}_\perp^t\| &= dt\ \ampli^{-\chi-1} \lsq-\frac{1}{\sqrt{d} \sqrt{2\pi}}+ \frac{T}{2\|\hat{\w}_\perp^t\|} c_n\rsq,
\end{aligned}
\label{eq:steady_highT_app}
\eeqn
and the asymptotic solution reads
\beqn
\begin{aligned}
\hat{w}_1^t &\sim k_1\ T \sqrt{d} \lpa \frac{t}{Td}\rpa^{\frac{1}{3+\chi}}\\
\|\hat{\w}^t_\perp\| &= k_\perp\ T \sqrt{d}
\end{aligned}
\label{eq:w_steady_app}
\eeqn
with the constants $k_\perp = c_n\sqrt{\frac{\pi}{2}}$ and $k_1 = k_\perp \lpa\frac{(\chi+3)c_1}{k_\perp}\rpa^{1/(\chi+3)}$.
Therefore, the ratio $\ampli=\hat{w}_1^t/\|\hat{\w}^t_\perp\|$ grows as a power law of time:
\beq
    \ampli \sim \frac{k_1}{k_\perp}\ \lpa\frac{t}{T d}\rpa^{\frac{1}{\chi+3}}.
    \label{eq:lambda_power}
\eeq
From this solution, using Eq. \ref{eq:n_asymp-app}, we obtain the asymptotic behavior of the fraction of unfitted points $\unf(\w^t)$:
\beqn
    \unf(\w^t) &\sim \ampli^{-\chi-1}
    \sim \lpa \frac{t}{Td}\rpa^{-\frac{\chi+1}{\chi+3}}.
\label{eq:n_wt_app}
\eeqn

\hypertarget{app:stoch_terms}{\subsection{Stochastic fluctuations in the high-temperature asymptotic solution}}
\label{app:stoch_terms}
To lighten the notation, we use $\hat{w}_\perp = \|\hat{\w}^t_\perp\|$.
In the limit $\ampli\rightarrow \infty$, $\aw\rightarrow 0$, from Eq. \ref{eq:steady_highT_app} we have
\beqn
g_1 &= \frac{c_1}{\sqrt{d}}\ampli^{-\chi-2}, \quad
\partial_{w_1} g_1 = -(\chi+2)\frac{c_1}{\sqrt{d}}\ampli^{-\chi-3} \frac{1}{\hat{w}_\perp}, \quad
\partial_{w_\perp} g_1 = (\chi+2)\frac{c_1}{\sqrt{d}}\ampli^{-\chi-3}\frac{w_1}{\hat{w}_\perp^2},
\eeqn

\beqn
\tilde{g}_\perp = \ampli^{-\chi-1}\lsq -\frac{1}{\sqrt{d 2\pi}}+\frac{c_n T}{2 w_\perp}\rsq, \quad
\partial_{w_1} \tilde{g}_\perp = 0, \quad
\partial_{w_\perp} \tilde{g}_\perp &= -\ampli^{-\chi-1} \frac{c_n T}{2 \hat{w}_\perp^2}.
\eeqn

Therefore, using Eq. \ref{eq:w_steady_app}, the matrices $\mathbf{A}$ and $\bm{\sigma}$ in Eq. \ref{eq:noise_dyn} are
\beq
\mathbf{A} = \frac{1}{k_\perp\ T d\ \ampli^{2+\chi}}
\lpa\begin{matrix}
-c_1 (\chi+2) \ampli^{-1} & c_1(\chi+2) \cr
0 & - \frac{\ampli}{\sqrt{2\pi}}
\end{matrix}\rpa
\eeq
and
\beq
\bm{\sigma} = \sqrt{\frac{T}{d}}\frac{1}{\ampli^{\frac{3+\chi}{2}}}
\lpa\begin{matrix}
\sqrt{c_{11}} & 0\cr
0 & \sqrt{c_{22}} \ampli
\end{matrix}\rpa.
\eeq
The formal solution to the time-dependent Ornstein–Uhlenbeck process Eq. \ref{eq:noise_dyn} is given by
\beq
    \mathbf{z}(t) = \int_{t_0}^t \mathbf{K}(t,s)\ \bm{\sigma}(s)\ d\tilde{\mathbf{W}}(s)
\eeq
with $\mathbf{K}(t,s) = \exp\lpa\int_s^t \mathbf{A}(t') dt'\rpa$, 
where, for large times, we are neglecting the initial condition $\mathbf{z}(t_0)$.
In the following, since $\ampli_t = \frac{\hat{w}_1}{\hat{w}_\perp} \sim (\frac{t}{Td})^{\frac{1}{3+\chi}}$ (Eq. \ref{eq:lambda_power}), we consider the beginning of the asymptotic solution at the time $t_0 \sim Td$ where $\ampli_{t_0}\sim 1$.\\


The eigenvalues of $\mathbf{A}$
are
\beq
\mu_1 = -a_1\ \frac{\ampli^{-\chi-3}}{Td}, \quad
\mu_2 = -a_2\ \frac{\ampli^{-\chi-1}}{Td},
\eeq
with pre-factors $a_1 = \frac{c_1(\chi+2)}{k_\perp}$, $a_2 = \frac{1}{\sqrt{2\pi} k_\perp}$, and the corresponding eigenvectors are
\beq
\mathbf{v}_1 = [1;0], \quad
\mathbf{v}_2 = [0; 1] + \O(\ampli^{-1}).
\eeq

Given the memory kernels
\begin{align}
\begin{aligned}
K_1(t,s) &= \exp\lpa \int_s^t \mu_1 dt' \rpa,\quad
K_2(t,s) = \exp\lpa \int_s^t \mu_2 dt' \rpa,
\end{aligned}
\end{align}

we compute the magnitude of $\langle z_1(t)^2\rangle$ and $\langle z_\perp(t)^2\rangle$ as 
\begin{align}
\begin{aligned}
\langle z_1(t)^2\rangle &= \int_{t_0}^t K_1(t,s)^2 \sigma_{11}^2 ds,\quad
\langle z_\perp(t)^2\rangle   &= \int_{t_0}^t K_2(t,s)^2 \sigma_{22}^2 ds.
\end{aligned}
\end{align}


To compute the first kernel, we use Eq. \ref{eq:lambda_power}, $\ampli = \frac{k_1}{k\perp} (\frac{t}{Td})^{1/(\chi+3)}$, to rewrite $\mu_1$ as $\mu_1 = -a_1\ \frac{\ampli^{-\chi-3}}{Td} = -\frac{\chi+2}{\chi+3} t^{-1} = -p \frac{1}{t}$ with $p=\frac{\chi+2}{\chi+3}<1$. Therefore:
\beq
K_1(t,s) = \exp\lpa \int_s^t \mu_1 dt' \rpa = \exp\lpa -p \int_s^t \frac{1}{t'}dt' \rpa =
\exp \ln{\lsq\lpa\frac{t}{s}\rpa^{-p}\rsq}
= \lpa \frac{t}{s}\rpa^{-p}.
\eeq

Since $\sigma_{11}^2 = \frac{T}{d}c_{11}\ampli^{-\chi-3} = \frac{c_{11}k_\perp}{(\chi+3)c_1} T^2 t^{-1}$, we have
\beq
\begin{aligned}
    \langle z_1(t)^2 \rangle &= \int_{t_0}^t K_1(t,s)^2 \sigma_{11}^2 ds = 
    \frac{c_{11}k_\perp}{(\chi+3)c_1} T^2 \int_{t_0}^t \lpa\frac{t}{s}\rpa^{-2p} s^{-1} ds
    \propto T^2 \lpa 1  - (t/t_0)^{-2p}\rpa.
\end{aligned}
\label{eq:z1}
\eeq

For the second kernel, we use Eq. \ref{eq:lambda_power} to rewrite $dt = \frac{k_\perp}{c_1} Td\ \ampli^{\chi+2}d\ampli$. Therefore:
\beq
    K_2(t,s) = \exp\lpa \int_s^t \mu_2 dt' \rpa = 
    \exp\lpa - \int_{\ampli_s}^{\ampli_t} \frac{\ampli^{-\chi-1}}{\sqrt{2\pi}k_\perp Td}\ \frac{k_\perp}{c_1} Td\ \ampli^{\chi+2}d\ampli \rpa=
    \exp\lpa -\frac{1}{\sqrt{2\pi}c_1} \int_{\ampli_s}^{\ampli_t} \ampli d\ampli \rpa=
    \exp\lpa -\frac{\ampli_t^2 - \ampli_s^2}{2\sqrt{2\pi} c_1} \rpa
\eeq

Since $\sigma_{22}^2=c_{22}\frac{T}{d} \ampli^{-\chi-1}$, we have
\begin{align}
\begin{aligned}
    \langle z_\perp(t)^2 \rangle &= \int_{t_0}^t K_2(t,s)^2 \sigma_{22}^2 ds
    = \int_{1}^{\ampli_t} \exp\lpa -\frac{\ampli_t^2 - \ampli_s^2}{\sqrt{2\pi} c_1} \rpa
    c_{22}\frac{T}{d} \ampli_s^{-\chi-1}\
    \frac{k_\perp}{c_1} Td\ \ampli_s^{\chi+2}d\ampli_s =\\
    &\propto T^2 \int_{1}^{\ampli_t} \exp\lpa -\frac{\ampli_t^2 - \ampli_s^2}{\sqrt{2\pi} c_1} \rpa
    \ampli_s d\ampli_s
    \sim T^2 \lpa 1 - e^{-\frac{1}{\sqrt{2\pi}c_1}\ampli_t^2}\rpa
\end{aligned}
\label{eq:zp}
\end{align}

Therefore, we obtain fluctuations $\langle z_1(t)^2 \rangle^{1/2} \propto T$ (Eq. \ref{eq:z1}), $\langle z_\perp(t)^2 \rangle^{1/2} \propto T$ (Eq. \ref{eq:zp}) which are negligible with respect to the deterministic solution $\hat{w}_1$, $\hat{w}_\perp$ for large $d$:
\beqn
\begin{aligned}
\frac{\langle z_1(t)^2 \rangle^{1/2}}{\hat{w}_1} &\sim d^{-1/2} \lpa\frac{t}{Td}\rpa^{-\frac{1}{1+\chi}},\\
\frac{\langle z_\perp(t)^2 \rangle^{1/2}}{\hat{w}_\perp} &\sim d^{-1/2}.
\end{aligned}
\eeqn

\hypertarget{app:break}{\section{Heuristic argument for the limit of validity of the online theory}}
\label{app:break}

The online SDE has an asymptotic solution that continues up to $t\rightarrow \infty$ since the number of training points $P$ diverges. For the empirical SGD dynamics, however, $P$ is finite and the dynamics stops at a time $t^*$ when all the training points have zero hinge loss, that is when $\unf(\w^{t^*}) = 0$. Therefore, we expect the existence of a cross-over time $\tcross < t^*$ when the finite-training-set fluctuations become important for the dynamics and the asymptotic solution of the online SDE is no longer a valid description of the empirical SGD.
To estimate the scaling of $\tcross$ with $P$ and $d$, we compare the population average of the loss gradient with its finite-training-set fluctuations. Assuming that the central limit theorem is valid for $\nabla L(\w^t)$, at leading order in $P$ we have
\beq
    -\nabla L(\w^t) = \mathbf{g}^t + \frac{1}{\sqrt{P}} \bm{\xi}^t,
\eeq
where $\bm{\xi}^t$ is a $d$-dimensional Gaussian vector with covariance $\cov^t$. In the asymptotic solution of the online dynamics, according to the expansions of Eq. \ref{eq:grad_asympt}, we have that
\beqn
\begin{aligned}
||\mathbf{g}^t|| &= \sqrt{(g_1^t)^2+(g_\perp^t)^2} = \frac{\ampli^{-\chi-1}}{\sqrt{2\pi d}} \lpa 1 + 2\pi \lpa\frac{c_1}{\ampli}\rpa^2\rpa^{1/2}
= \frac{\ampli^{-\chi-1}}{\sqrt{2\pi d}} \lpa 1+\O(\ampli^{-2})\rpa
=\frac{n(\w^t)}{\sqrt{d}\ c_n \sqrt{2\pi}} + \O\lpa\frac{\ampli^{-2}}{\sqrt{d}}\rpa.
\end{aligned}
\label{eq:norm_gt}
\eeqn
Using the covariance expression Eq. \ref{eq:cov_app}:
\beqn
\begin{aligned}
||\bm{\xi}^t|| &= \Tr\lpa\cov^t\rpa^{1/2} = \Tr\lpa\frac{\unf(\w^t)}{d}\tilde{\I}_{d-2}  + \frac{\tilde{\cov}}{d}\rpa^{1/2}
=\lpa \unf(\w^t) \frac{d-2}{d} + \frac{\Tr\tilde{\cov}}{d}\rpa^{1/2}
=\sqrt{\unf(\w^t)} \lpa 1 + \frac{\Tr\tilde{\cov}}{(d-2)\unf(\w^t)}\rpa^{1/2}=\\
&=\sqrt{\unf(\w^t)} + \O(d^{-1})
\end{aligned}
\label{eq:norm_xi}
\eeqn

We can neglect the finite $P$ fluctuations as long as $\|\mathbf{g}^t\|\gg \frac{1}{\sqrt{P}}\|\bm{\xi}^t\|$. From Eq. \ref{eq:norm_gt} and Eq. \ref{eq:norm_xi}, this corresponds to $\unf(\w^t)\gg 2\pi c_n^2\ d/P$. Defining $\tcross$ as the time for which $\|\mathbf{g}^\tcross\| \sim \frac{1}{\sqrt{P}}\|\mathbf{\bm{\xi}}^\tcross\|$, we have
\beq
\tcross: \quad \unf (\w^\tcross ) \sim \O\lpa\frac{d}{P}\rpa.
\label{eq:perceptron_cross}
\eeq

Using the asymptotic behaviour Eq. \ref{eq:n_wt_app}, we obtain the scaling of the cross-over time $\tcross$ as
\beq
\tcross \sim d^{1-b}\ T P^b,
\label{eq:tcross_scaling}
\eeq
with $b=1+\frac{2}{\chi+1}$. From Eq. \ref{eq:w_steady_app}, we obtain the scaling of the informative weight $w_1^{\tcross}$ as
\beq
    w_1^{\tcross} \sim d^{\frac{1}{2}-\ec}\ T P^{\ec},
    \label{eq:wcross_scaling}
\eeq
with $\ec = \frac{1}{1+\chi}$.

\hypertarget{app:plotsPerc}{\subsection{Empirical test of the heuristic argument}}
\label{app:plotsPerc}
\quad 
\begin{figure}[H]
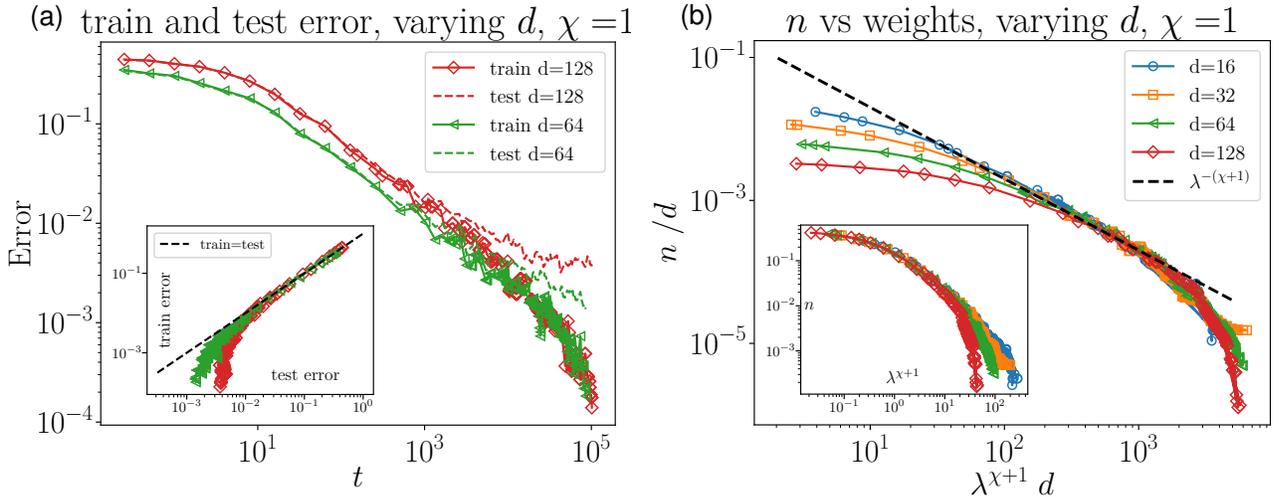

    \centering
    \begin{tikzpicture}
        \node[anchor=north west,inner sep=0pt] at (0,0){
        \resizebox{.48\textwidth}{!}{
        \input{figures/depleted_sign-train_test_error-single-d-chi1-P8192-T0125.pgf}}};
        \node[font=\sffamily\Large, fill=white!20, minimum size=0.8cm] at (5.7ex,-2.ex) {(a)};
    \end{tikzpicture}
    \begin{tikzpicture}
        \node[anchor=north west,inner sep=0pt] at (0,0){
        \resizebox{.48\textwidth}{!}{
        \input{figures/depleted_sign-n_weights-new-d-chi1-P4096-T0125.pgf}}};
        \node[font=\sffamily\Large, fill=white!20, minimum size=0.8cm] at (5.7ex,-2.ex) {(b)};
    \end{tikzpicture}
    \caption{\textbf{SGD is well described by its online approximation up to a time $\tcross$ depending on the dimension $d$.}
    \textbf{(a) Evolution of the training and test errors in time $t$, varying $d$, fixed $P=8192$.} \textit{Main panel.} The training and test errors are on top of each other at small $t$ and separate later on in the dynamics at a time $\tcross$ depending on $d$.
    \textit{Inset:} Plotting train error vs the test error during the training dynamics the two quantities are equal (diagonal black dashed line) and deviate from the diagonal only at small values of the training error, when the latter converges to zero while the test error plateaus to a finite value depending on $d$.
    \textbf{(b) Fraction of points with non-zero loss $n(\w^t)$ as a function of $\ampli^{\chi+1} = \lpa\frac{w_1^t}{\|\w_\perp^t\|}\rpa^{\chi+1}$, varying $d$, fixed $P=4096$.
    } 
    \textit{Inset:} $n(\w^t)$ decreases as a power law of $\ampli$ up to a cross-over point depending on $d$.
    \textit{Main panel:} $n(\w^t)$ decreases as $n(\w^t)\sim \ampli^{-(\chi+1)}$, accordingly to the online asymptotic solution Eq. \ref{eq:n_asymp-app} (black dashed lines). Rescaling the axis by $d$ aligns the cross-over points and the curves at the end of training, consistently with Eq. \ref{eq:perceptron_cross} $\unf(\w^\tcross)\sim d$.
    }
    \label{fig:early_stopping}
\end{figure}

\begin{figure}[H]
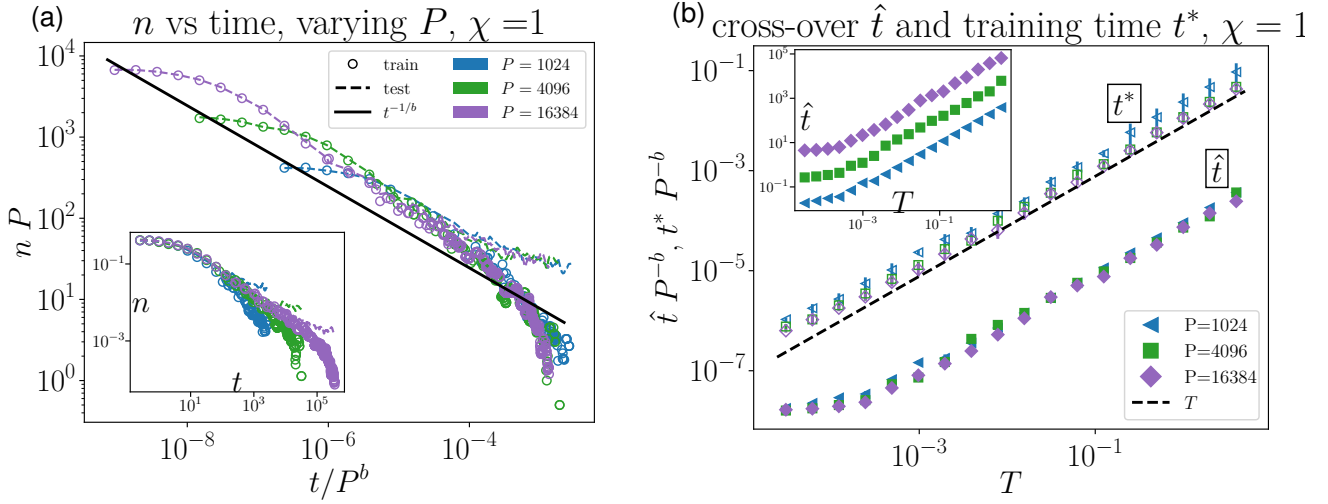

    \centering
    \begin{tikzpicture}
        \node[anchor=north west,inner sep=0pt] at (0,0){
        \resizebox{.46\textwidth}{!}{
        \input{figures/depleted_sign-n_test_time-single-P-chi1-d128-T0125.pgf}}};
        \node[font=\sffamily\Large] at (5.7ex,-2.ex) {(a)};
    \end{tikzpicture}
    \hspace{.2cm}
    \begin{tikzpicture}
        \node[anchor=north west,inner sep=0pt] at (0,0){
        \resizebox{.5\textwidth}{!}{
        \input{figures/depleted_sign-cross_time-single_TP-chi1-d128.pgf}}};
        \node[font=\sffamily\Large] at (5.2ex,-2.ex) {(b)};
    \end{tikzpicture}
    \caption{
    \textbf{The scaling argument Eq. \ref{eq:perceptron_cross} for $\tcross$ captures also the $T$ and $P$ dependence of the final time $\tf$.}
    \textbf{(a) Fraction of points with non-zero loss $n(\w^t)$ as a function of time $t$, varying $P$ and fixed $d=128$.
    } 
    \textit{Inset:} $n(\w^t)$ reaches zero at the end of training when computed on the training set, while it plateaus to a finite value for the test set. The two curves (train and test $n(\w^t)$) separate at a cross-over time $\tcross$ depending on $P$. \textit{Main panel:} Rescaling the $y$-axis by $P$ and and the $x$-axis by $P^{-b}$, $b=1+\frac{2}{\chi+1}$, aligns the points where the two curves separate, consistently with Eq. \ref{eq:perceptron_cross} $n(\w^{\tcross})\sim P^{-1}$ and Eq. \ref{eq:tcross_scaling} $\tcross \sim P^b$.
    This rescaling collapses training and test curves also at the end of training, indicating that $\tcross$ and the total training time $t^*$ have the same $P$ dependence.
    The black line represents the online-SDE asymptotic solution Eq. \ref{eq:n_wt_app} $n(\w^t)\sim t^{-\frac{\chi+1}{\chi+3}} = t^{-1/b}$.
    \textbf{(b) Cross-over $\tcross$ (full symbols) and training time $\tf$ (empty symbols) vs SGD temperature $T$, varying training set size $P$, ($\chi=1$, $d=128$).} \textit{Inset:} $\tcross$ grows both with $T$ and $P$. \textit{Main:} $\tcross$ grows proportional to $T$ and data for different $P$ are collapsed by the rescaling $P^{-b}$ with $b=1+2/(1+\chi)$, consistently with \ref{eq:tcross_scaling}. $\tf$ is found to have the same scaling in $P$ and $T$.
    }
    \label{fig:n_w_t}
\end{figure}

\begin{figure}[H]
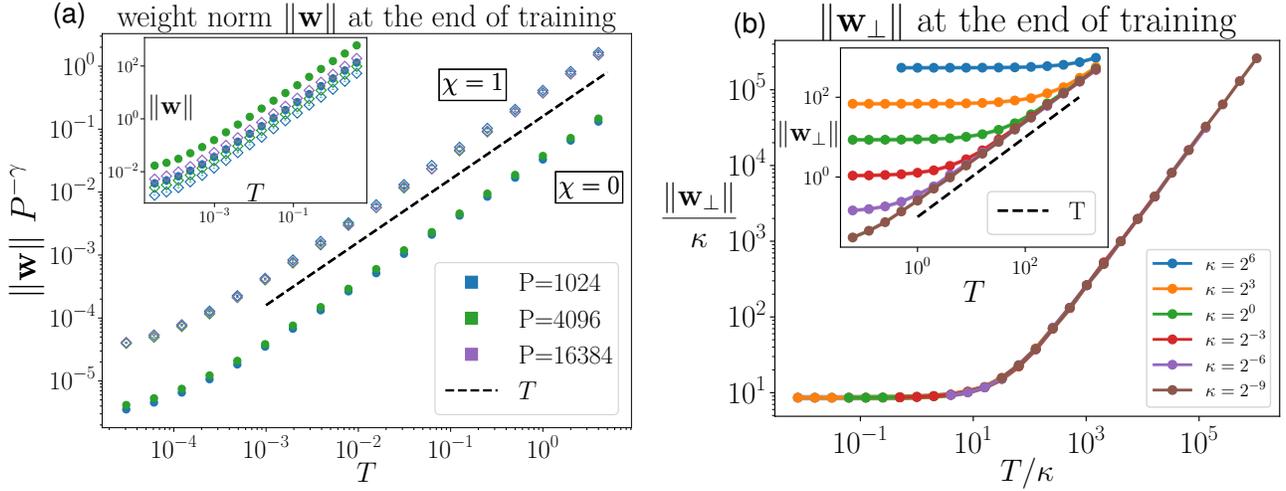

    \centering
    \begin{tikzpicture}
        \node[anchor=north west,inner sep=0pt] at (0,0){
        \resizebox{.48\textwidth}{!}{
        \input{figures/depleted_sign-w_variation-single_TPchi-d128.pgf}}};
        \node[font=\sffamily\Large, fill=white!20, minimum size=0.8cm] at (7ex,-2.ex) {(a)};
    \end{tikzpicture}
    \begin{tikzpicture}
        \node[anchor=north west,inner sep=0pt] at (0,0){
        \resizebox{.48\textwidth}{!}{
        \input{figures/depleted_sign-w_perp_margin-chi1.0-d128-P8192.pgf}}};
        \node[font=\sffamily\Large] at (10ex,-3.ex) {(b)};
    \end{tikzpicture}
    \caption{
    \textbf{(a) The scaling argument Eqs. \ref{eq:perceptron_cross}-\ref{eq:wcross_scaling} predicts the $T$ and $P$ dependence of the norm of the perceptron weights $\|\w\|$ at the end of training.}
    Data for fixed $d=128$. Empty (full) symbols correspond to data distribution $\chi=1$ ($\chi=0$). \textit{Inset:} $\|\w\|$ grows both with $T$ and $P$. \textit{Main:} $\|\w\|$ grows proportional to $T$ and data for different $P$ are collapsed by the rescaling $P^{-\ec}$ with $\ec=1/(1+\chi)$, consistently with Eq. \ref{eq:wcross_scaling}.
    \textbf{(b) $\|\w_\perp\|$ at the end of training grows with $T$ when $T\gg \ma$.}
    \textit{Inset:} $\|\w_\perp\|$ grows proportionally to $T$ when $T$ is larger than a $\ma$-dependent value. \textit{Main:} Data are collapsed by rescaling both axis by $\ma$, suggesting that $\|\w_\perp\|\propto \ma$ when $T\ll \ma$ and $\|\w_\perp\|\propto T$ when $T\gg \ma$.
    Data for $d=128$, $P=8192$, $\chi=1$.
    }
    \label{fig:wnorm}
\end{figure}

\hypertarget{app:sde}{\section{Stochastic differential equation}}
\label{app:sde}

\hypertarget{app:summary}{\subsection{Summary: quantities averaged over the data distribution}}
\label{app:summary}
We report the expressions of $\mathbf{g}^t = \E_{\x}\lsq -\nabla L(\w^t)\rsq$, $\cov^t = \E_{\x}\lsq \nabla L(\w^t) \otimes \nabla L(\w^t)\rsq - {\mathbf{g}^t}\otimes {\mathbf{g}^t}$ and $\unf^t = \E_{\x}\lsq\theta\lpa\ma-y(\x)f(\w^t,\x)\rpa\rsq$ computed in \ref{app:depletion}. We remind that 
\begin{itemize}
    \item the direction $\ewt$ is the teacher direction;
    \item $\w=w_1\ewt + \w_\perp$;
    \item $\ewp = \w_\perp / \|\w_\perp\|$;
    \item $\rho(x_1) = x_1^\chi e^{-x_1^2/2}/Z$;   $Z= 2^{\frac{1+\chi}{2}}\Gamma(\frac{1+\chi}{2})$.
\end{itemize}
All the quantities are expressed as functions of the time-dependent ratios
\beq
    \ampli = \frac{w_1^t}{\|\w_\perp^t\|},\quad
    \aw = \frac{\ma \sqrt{d}}{\|\w_\perp^t\|}.
\eeq

\paragraph{Gradient terms.}
We use the notation
\beq
g_1 = \E_{\x}\lsq -\partial_{w_1} L(\w^t)\rsq, \quad
\mathbf{g}_\perp = \E_{\x}\lsq -\partial_{\w_\perp} L(\w^t)\rsq.
\eeq

\beqn
\begin{aligned}
g_1(w_1^t, \|\w_\perp^t\|) &= \frac{1}{\sqrt{d}} \int_0^{\infty} dx_1 \rho(x_1) x_1 \lsq 1+ \erf\lpa\frac{\aw-\ampli x_1}{\sqrt{2}}\rpa \rsq\\
\end{aligned}
\eeqn

\beqn
\begin{aligned}
\mathbf{g}_\perp(w_1^t, \|\w_\perp^t\|) &= -\frac{2}{\sqrt{d} \sqrt{2\pi}} \int_0^{\infty} dx_1 \rho(x_1) e^{-\frac{\lpa\aw-\ampli x_1\rpa^2}{2}}\ \ewp\\
g_\perp(w_1^t, \|\w_\perp^t\|) &= \ewp \cdot \mathbf{g}_\perp(w_1^t, \|\w_\perp^t\|)\\
\end{aligned}
\eeqn

\beqn
\begin{aligned}
\unf(w_1^t, \|\w_\perp^t\|) & = \int_0^{\infty} dx_1 \rho(x_1) \lsq 1+ \erf\lpa\frac{\aw-\ampli x_1}{\sqrt{2}}\rpa \rsq\\
\end{aligned}
\eeqn

For $\ampli\rightarrow\infty, \aw\rightarrow 0$:
\begin{align}
    \label{eq:g1_asymp}
    g_1  = &\hspace{0.3cm} \frac{c_1}{\sqrt{d}} \ampli^{-\chi-2} + \O\lpa\ampli^{-\chi-4}\rpa,\quad 
    \text{with  } c_1 = \frac{1+\chi}{\sqrt{2\pi}(2+\chi)} + \O(\aw),\\
    \label{eq:gp_asymp}
    g_\perp = &-\frac{c_2}{\sqrt{d}} \ampli^{-\chi-1} + \O\lpa\ampli^{-\chi-3}\rpa,\quad 
    \text{with  } c_2 = \frac{1}{\sqrt{2\pi}} + \O(\aw),\\
    \label{eq:n_asymp-app}
    \unf  = &\hspace{0.3cm} c_\unf  \ampli^{-\chi-1} + O\lpa \ampli^{-\chi-3}\rpa,\quad 
    \text{with  } c_\unf = \frac{\Gamma(1+\frac{\chi}{2})}{2\sqrt{\pi}\Gamma(\frac{3+\chi}{2})} + \O(\aw).
\end{align}

\paragraph{Covariance terms.}
\beqn
\cov(w_1^t, \|\w_\perp^t\|) & = \frac{1}{d}\tilde{\cov}^t + \frac{\unf^t}{d}\ \tilde{\I}_{d-2},
\label{eq:cov_app}
\eeqn
\beqn
\sqrt{\cov^t} = \sqrt{\frac{\unf^t}{d}}\ \tilde{\I}_{d-2} + \sqrt{\frac{\tilde{\cov}^t}{d}},
\eeqn

with
\beqn
\tilde{\cov}^t = \tilde{\Sigma}_{11}^t \ewt \otimes \ewt + \tilde{\Sigma}_{12}^t \lpa\ewt \otimes \ewp + \ewp \otimes \ewt \rpa +\tilde{\Sigma}_{22}^t \ewp \otimes \ewp,
\eeqn
\beq
\tilde{\I}_{d-2} = \I_d - \ewt \otimes \ewt - \ewp \otimes \ewp,
\eeq

where $\I_d$ is the $d\times d$ identity matrix and 
\beqn
\tilde{\Sigma}_{11}^t &= \int_0^{\infty} dx_1 \rho(x_1) x_1^2 \lpa 1+ \erf\lpa\frac{\aw-\ampli x_1}{\sqrt{2}}\rpa \rpa 
- \lsq\int_0^{\infty} dx_1 \rho(x_1) x_1 \lpa 1+ \erf\lpa\frac{\aw-\ampli x_1}{\sqrt{2}}\rpa \rpa\rsq^2,
\eeqn
\beqn
\tilde{\Sigma}_{12}^t = - \frac{2}{\sqrt{2\pi}}\ \int_0^{\infty} dx_1 \rho(x_1) x_1 e^{-\frac{\lpa\aw-\ampli x_1\rpa^2}{2}}
+\frac{2}{\sqrt{2\pi}} \lsq\int_0^{\infty} dx_1 \rho(x_1) x_1 \lpa 1+ \erf\lpa\frac{\aw-\ampli x_1}{\sqrt{2}}\rpa \rpa\rsq
\int_0^{\infty} dx_1 \rho(x_1) e^{-\frac{\lpa\aw-\ampli x_1\rpa^2}{2}},
\eeqn
\beqn
\tilde{\Sigma}_{22}^t = \unf^t -\frac{2}{\sqrt{2\pi}} \int_0^{\infty} dx_1 \rho(x_1) e^{-\frac{\lpa\aw-\ampli x_1\rpa^2}{2}}\lpa\aw-\ampli x_1\rpa
- \lsq\frac{2}{\sqrt{2\pi}} \int_0^{\infty} dx_1 \rho(x_1) e^{-\frac{\lpa\aw-\ampli x_1\rpa^2}{2}}\rsq^2.
\eeqn

We further define:
\beqn
\tilde{\sigma}_{11} = \lpa\sqrt{\tilde{\cov}}\rpa_{11},\quad
\tilde{\sigma}_{12} = \lpa\sqrt{\tilde{\cov}}\rpa_{12},\quad
\tilde{\sigma}_{22} = \lpa\sqrt{\tilde{\cov}}\rpa_{22},
\eeqn


\beqn
\sigma_1 = \sqrt{\tilde{\sigma}_{11}^2+\tilde{\sigma}_{12}^2} = \lpa\tilde{\Sigma}_{11}\rpa^{1/2},
\eeqn
\beqn
\sigma_2 = \sqrt{\tilde{\sigma}_{22}^2+\tilde{\sigma}_{12}^2} = \lpa\tilde{\Sigma}_{22}\rpa^{1/2}.
\eeqn

In the asymptotic limit $\ampli\rightarrow\infty$, $\aw\rightarrow 0$:
\beqn
\tilde{\Sigma}_{11}^t &= \lsq\frac{c_{11}}{\ampli^{\chi+3}} +\O\lpa\ampli^{-(\chi+5)}\rpa\rsq- \lsq\frac{c_{1}^2}{\ampli^{2\lpa\chi+2\rpa}} + \O\lpa\lambda^{-2(\chi+3)}\rpa\rsq
= c_{11} \ampli^{-\chi-3} + \O\lpa\ampli^{-min(\chi+5, 2\chi+4)}\rpa,
\label{eq:Sigma_11}
\eeqn
\beqn
\begin{aligned}
    \tilde{\Sigma}_{22}^t &= 
    \lsq \frac{c_{n}}{\ampli^{\chi+1}} - \frac{\tilde{c}_{22}}{\ampli^{\chi+1}} + \O\lpa\ampli^{-\chi-3}\rpa\rsq 
    -\lsq \frac{c_2^2}{\ampli^{2\chi+2}} + \O\lpa\ampli^{-2\chi-4}\rpa\rsq 
    = c_{22} \ampli^{-\chi-1} + \O\lpa\ampli^{-min(\chi+3, 2\chi+2)}\rpa,
\end{aligned}
\label{eq:Sigma_pp}
\eeqn

with $c_{11} = \frac{2\ \Gamma(2+\frac{\chi}{2})}{\sqrt{\pi}(3+\chi)\Gamma(\frac{1+\chi}{2})} + \O(\aw)$,\ 
$c_{22} = c_{n} - \tilde{c}_{22} = \frac{2+\chi}{2\sqrt{\pi} \Gamma(\frac{3+\chi}{2})} + \O(\aw)$,
from which we get

\beq
\sigma_1 \sim \sqrt{c_{11}} \ampli^{-\frac{\chi+3}{2}},
\eeq
\beq
\sigma_2 \sim \sqrt{c_{22}} \ampli^{-\frac{\chi+1}{2}}.
\eeq

\hypertarget{app:onlineSDE_w1}{\subsection{Online SDE for $w_1$ and $\|\w_\perp\|$}}
\label{app:onlineSDE_w1}
Given the online SDE
\beq
d\w^t = dt \mathbf{g}^t + T \sqrt{\cov^t} d\mathbf{W}^t,
\label{eq:onlineSDE}
\eeq
the equation for $w_1$ is obtained by projecting Eq. \ref{eq:onlineSDE} on $\ewt$:
\beq
    dw_1^t = g_1^t dt + \sqrt{\frac{T}{d}} \lpa \tilde{\sigma}_{11} dW_1^t + \tilde{\sigma}_{12} dW_2^t\rpa =
    g_1^t dt + \sqrt{\frac{T}{d}} \sigma_1 d\tilde{W}_1^t,
\label{eq:w1_sde-app}
\eeq
where we have used the substitution $\tilde{\sigma}_{11} dW_1^t + \tilde{\sigma}_{12} dW_2^t = \sigma_1 d\tilde{W}_1^t$ with $\sigma_1 = \sqrt{\tilde{\sigma}_{11}^2+\tilde{\sigma}_{12}^2}$.
$\tilde{W}_1^t$ is a Brownian motion correlated to $W_1^t$ and $W_2^t$.\\

The equation for $\|\w_\perp\|=f(\w_\perp)$ is obtained by considering Eq. \ref{eq:onlineSDE} for $d \w_\perp^t$, 
\beq
    d \w_\perp^t = \mathbf{g}_\perp^t dt + \sqrt{T}\ \tilde{\I}_{d-1} \sqrt{\cov^t} d\mathbf{W}^t,
\eeq
where $\tilde{\I}_{d-1} = \I_d-\ewt\otimes\ewt$ is the projector in the directions orthogonal to $\ewt$, and applying Ito's lemma:
\beq
    d \|\w_\perp^t\| = \lsq
    \frac{\partial f}{\partial \w_\perp} \cdot \mathbf{g}_\perp^t +  
    +\frac{1}{2} \Tr\lsq (\tilde{\I}_{d-1}\sqrt{T\cov^t})^T \cdot \frac{\partial^2 f}{\partial \w_\perp \partial \w_\perp} \cdot (\tilde{\I}_{d-1}\sqrt{T\cov^t})\rsq
    \rsq dt + 
    \frac{\partial f}{\partial \w_\perp} \cdot (\tilde{\I}_{d-1}\sqrt{T}\sqrt{\cov^t})d\mathbf{W}^t.
\eeq

Using the gradient
\beq
    \frac{\partial f}{\partial \w_\perp} = \frac{\partial \|\w_\perp\|}{\partial \w_\perp} = \ewp,
\eeq
and the Hessian
\beq
    \frac{\partial^2 f}{\partial \w_\perp \partial \w_\perp} = \frac{\partial^2 \|\w_\perp\|}{\partial \w_\perp \partial \w_\perp} = \frac{1}{\|\w_\perp\|}\lpa \tilde{\I}_{d-1} - \ewp\otimes\ewp\rpa,
\eeq
we get:
\beq
    \frac{\partial f}{\partial \w_\perp} \cdot \mathbf{g}_\perp^t = g_\perp^t,
\eeq
\beqn
\begin{aligned}
&\Tr\lsq (\tilde{\I}_{d-1}\sqrt{T\cov^t})^T \cdot \frac{\partial^2 f}{\partial \w_\perp \partial \w_\perp} \cdot (\tilde{\I}_{d-1}\sqrt{T\cov^t})\rsq
= \frac{T}{\|\w_\perp\|} \Tr\lsq\lpa \tilde{\I}_{d-1} - \ewp\otimes\ewp\rpa \cov\rsq = \\
&=\frac{T}{\|\w_\perp\|} \Tr\lsq \frac{\unf}{d} \tilde{\I}_{d-2}\rsq
= \frac{T}{\|\w_\perp\|} \frac{d-2}{d}\ \unf \simeq \frac{T}{\|\w_\perp\|} \unf \quad \textrm{for }d\gg1,
\end{aligned}
\eeqn
\beqn
    \frac{\partial f}{\partial \w_\perp} \cdot (\tilde{\I}_{d-1}\sqrt{T}\sqrt{\cov})d\mathbf{W} &= \sqrt{\frac{T}{d}}\lpa \tilde{\sigma}_{12} dW_1 + \tilde{\sigma}_{22}dW_2\rpa = 
    \sqrt{\frac{T}{d}} \sigma_2 d\tilde{W}_2,
\eeqn
where we have used the substitution $\tilde{\sigma}_{12} dW_1^t + \tilde{\sigma}_{22} dW_2^t = \sigma_2 d\tilde{W}_2^t$ with $\sigma_2 = \sqrt{\tilde{\sigma}_{12}^2+\tilde{\sigma}_{22}^2}$.
The Brownian motion $\tilde{W}_2$ is correlated to $W_1$ and $W_2$, and therefore is correlated also to $\tilde{W}_1$ appearing in the SDE of $w_1$ (Eq. \ref{eq:w1_sde-app}).
We obtain
\beq
d\|\w_\perp^t\| = \lsq g_\perp^t + \frac{T}{2 \|\w_\perp^t\|}\unf^t  \rsq dt + \sqrt{\frac{T}{d}} \sigma_2^t d\tilde{W}_2^t.
\eeq

\hypertarget{app:depletion}{\subsection{Computations of the averages}}
\label{app:depletion}
We consider a data distribution that is standard Gaussian on $\x_\perp$ and varies on $x_1$ as $\rho(x_1) = \frac{x_1^\chi e^{-x_1^2/2}}{2^{\frac{1+\chi}{2}}\Gamma(\frac{1+\chi}{2})}$.

\paragraph{Computation of } $g_1$

\beqn
\begin{aligned}
    & \sqrt{d} g_1 = \E_{\x\sim \rho} \lsq x_1\ \textrm{sign}(x_{1}) \theta\lpa \ma -\textrm{sign}(x_{1}) w_1 x_1/\sqrt{d} - \textrm{sign}(x_{1}) \w_\perp \cdot \x_\perp/\sqrt{d} \rpa
\rsq=\\
    &=\int_0^{\infty} dx_1 \rho(x_1) x_1 \int d\x_{\perp} \rho(\x_{\perp}) \theta\lpa \sqrt{d}\ma - w_1 x_1 - \w_\perp \cdot \x_\perp \rpa 
    +\int_{-\infty}^{0} dx_1 \rho(x_1) (-x_1) \int d\x_{\perp} \rho(\x_{\perp}) \theta\lpa \sqrt{d}\ma + w_1 x_1 + \w_\perp \cdot \x_\perp \rpa =\\
    &=\int_0^{\infty} dx_1 \rho(x_1) x_1 \int_{-\infty}^{\infty} dz \rho(z) \theta\lpa \sqrt{d}\ma - w_1 x_1 - \|\w_\perp\| z \rpa +
    \int_0^{\infty} dx_1 \rho(x_1) x_1 \int_{-\infty}^{\infty} dz \rho(z) \theta\lpa \sqrt{d}\ma - w_1 x_1 + \|\w_\perp\| z \rpa =\\
    &=\int_0^{\infty} dx_1 \rho(x_1) x_1 \lsq 1+ \erf\lpa\frac{\aw-\ampli x_1}{\sqrt{2}}\rpa \rsq
\end{aligned}
\eeqn

\paragraph{Computation of } $g_\perp$

\begin{align}
\begin{aligned}
    \sqrt{d} \mathbf{g}_\perp  &= \E_{\x\sim \rho} \lsq \x_\perp \ \sign(x_{1}) \theta\lpa \ma - \sign(x_{1}) w_1 x_1/\sqrt{d} - \sign(x_{1}) \w_\perp \cdot \x_\perp/\sqrt{d} \rpa \rsq=\\
    &=\int_0^{\infty} dx_1 \rho(x_1) \int d\x_{\perp} \rho(\x_{\perp}) \x_\perp \theta\lpa \sqrt{d}\ma - w_1 x_1 - \w_\perp \cdot \x_\perp \rpa
    -\int_{-\infty}^{0} dx_1 \rho(x_1) \int d\x_{\perp} \rho(\x_{\perp}) \x_\perp \theta\lpa \sqrt{d}\ma + w_1 x_1 + \w_\perp \cdot \x_\perp \rpa =\\
    &=\int_0^{\infty} dx_1 \rho(x_1)  \int_{-\infty}^{\infty} dz \rho(z) z\ \theta\lpa \sqrt{d}\ma - w_1 x_1 - \|\w_\perp\| z \rpa \ewp
    -\int_0^{\infty} dx_1 \rho(x_1)  \int_{-\infty}^{\infty} dz \rho(z) z\ \theta\lpa \sqrt{d}\ma - w_1 x_1 + \|\w_\perp\| z \rpa \ewp=\\
    &=-\frac{2}{\sqrt{2\pi}} \int_0^{\infty} dx_1 \rho(x_1) e^{-\frac{\lpa\aw-\ampli x_1\rpa^2}{2}} \ewp
\end{aligned}
\end{align}

\paragraph{Computation of } $\unf$
\beqn
\begin{aligned}
    \unf &= \E_{\x\sim \rho} \lsq \theta\lpa \ma - \textrm{sign}(x_{1}) w_1 x_1/\sqrt{d} - \textrm{sign}(x_{1}) \w_\perp \cdot \x_\perp/\sqrt{d} \rpa
\rsq=\\
    &=\int_0^{\infty} dx_1 \rho(x_1) \int d\x_{\perp} \rho(\x_{\perp}) \theta\lpa \sqrt{d}\ma - w_1 x_1 - \w_\perp \cdot \x_\perp \rpa
    +\int_{-\infty}^{0} dx_1 \rho(x_1) \int d\x_{\perp} \rho(\x_{\perp}) \theta\lpa \sqrt{d}\ma + w_1 x_1 + \w_\perp \cdot \x_\perp \rpa =\\
    &= \int_0^{\infty} dx_1 \rho(x_1)  \int_{-\infty}^{\infty} dz \rho(z)\ \theta\lpa \sqrt{d}\ma - w_1 x_1 - \|\w_\perp\| z \rpa
    +\int_0^{\infty} dx_1 \rho(x_1)  \int_{-\infty}^{\infty} dz \rho(z)\ \theta\lpa \sqrt{d}\ma - w_1 x_1 + \|\w_\perp\| z \rpa=\\
    &= \int_0^{\infty} dx_1 \rho(x_1) \lsq 1+ \erf\lpa\frac{\aw-\ampli x_1}{\sqrt{2}}\rpa \rsq
\end{aligned}
\eeqn

\paragraph{Computation of } $C_{11}= \E_{\x\sim \rho}\lsq x_1^2 \theta\lpa\sqrt{d}\ma - y(\x) \w\cdot\x \rpa\rsq$
\beqn
\begin{aligned}
    &\E_{\x\sim \rho} \lsq x_1^2 \theta\lpa \sqrt{d}\ma - \textrm{sign}(x_{1}) w_1 x_1 - \textrm{sign}(x_{1}) \w_\perp \cdot \x_\perp \rpa
\rsq=\\
    &=\int_0^{\infty} dx_1 \rho(x_1) x_1^2 \int d\x_{\perp} \rho(\x_{\perp}) \theta\lpa \sqrt{d}\ma - w_1 x_1 - \w_\perp \cdot \x_\perp \rpa
    +\int_{-\infty}^{0} dx_1 \rho(x_1) x_1^2 \int d\x_{\perp} \rho(\x_{\perp}) \theta\lpa \sqrt{d}\ma + w_1 x_1 + \w_\perp \cdot \x_\perp \rpa =\\
    &= \int_0^{\infty} dx_1 \rho(x_1) x_1^2 \int_{-\infty}^{\infty} dz \rho(z)\ \theta\lpa \sqrt{d}\ma - w_1 x_1 - \|\w_\perp\| z \rpa
    +\int_0^{\infty} dx_1 \rho(x_1) x_1^2 \int_{-\infty}^{\infty} dz \rho(z)\ \theta\lpa \sqrt{d}\ma - w_1 x_1 + \|\w_\perp\| z \rpa=\\
    &= \int_0^{\infty} dx_1 \rho(x_1) x_1^2 \lsq 1+ \erf\lpa\frac{\aw-\ampli x_1}{\sqrt{2}}\rpa \rsq
\end{aligned}
\eeqn

\paragraph{Computation of } $C_{1\perp}= \E_{\x\sim \rho}\lsq x_1 \x_\perp \theta\lpa\sqrt{d}\ma - y(\x) \w\cdot\x \rpa\rsq$
\beqn
\begin{aligned}
     &\E_{\x\sim \rho} \lsq x_1 \x_\perp \theta\lpa \sqrt{d}\ma - \textrm{sign}(x_{1}) w_1 x_1 - \textrm{sign}(x_{1}) \w_\perp \cdot \x_\perp \rpa \rsq=\\
    &=\int_0^{\infty} dx_1 \rho(x_1) x_1 \int d\x_{\perp} \rho(\x_{\perp}) \x_\perp \theta\lpa \sqrt{d}\ma - w_1 x_1 - \w_\perp \cdot \x_\perp \rpa + \\
    &+\int_{-\infty}^{0} dx_1 \rho(x_1) x_1 \int d\x_{\perp} \rho(\x_{\perp})\x_\perp \theta\lpa \sqrt{d}\ma + w_1 x_1 + \w_\perp \cdot \x_\perp \rpa =\\
    &= \int_0^{\infty} dx_1 \rho(x_1) x_1 \int_{-\infty}^{\infty} dz \rho(z)\ z \theta\lpa \sqrt{d}\ma - w_1 x_1 - \|\w_\perp\| z \rpa\ \ewp +\\
    &-\int_0^{\infty} dx_1 \rho(x_1) x_1 \int_{-\infty}^{\infty} dz \rho(z)\ z \theta\lpa \sqrt{d}\ma - w_1 x_1 + \|\w_\perp\| z \rpa\ \ewp =\\
    &= - \frac{2}{\sqrt{2\pi}}\ \int_0^{\infty} dx_1 \rho(x_1) x_1 e^{-\frac{\lpa\aw-\ampli x_1\rpa^2}{2}}\ \ewp
\end{aligned}
\eeqn

\paragraph{Computation of } $C_{\perp\perp}= \E_{\x\sim \rho}\lsq \x_\perp \otimes \x_\perp \theta\lpa\sqrt{d}\ma - y(\x) \w\cdot\x \rpa\rsq$
\beqn
\begin{aligned}
    &\E_{\x\sim \rho} \lsq \x_\perp \otimes \x_\perp \theta\lpa \sqrt{d}\ma - \textrm{sign}(x_{1}) w_1 x_1 - \textrm{sign}(x_{1}) \w_\perp \cdot \x_\perp \rpa
\rsq=\\
    &=\int_0^{\infty} dx_1 \rho(x_1) \int d\x_{\perp} \rho(\x_{\perp}) \x_\perp \otimes \x_\perp \theta\lpa \sqrt{d}\ma - w_1 x_1 - \w_\perp \cdot \x_\perp \rpa + \\
    &+\int_{-\infty}^{0} dx_1 \rho(x_1) \int d\x_{\perp} \rho(\x_{\perp}) \x_\perp \otimes \x_\perp \theta\lpa \sqrt{d}\ma + w_1 x_1 + \w_\perp \cdot \x_\perp \rpa =\\
    &= \int_0^{\infty} dx_1 \rho(x_1) \int_{-\infty}^{\infty} dz \rho(z)\ z^2 \theta\lpa \sqrt{d}\ma - w_1 x_1 - \|\w_\perp\| z \rpa\ \ewp\ewp^T +\\
    &+\int_0^{\infty} dx_1 \rho(x_1) \int_{-\infty}^{\infty} dz \rho(z)\ z^2 \theta\lpa \sqrt{d}\ma - w_1 x_1 + \|\w_\perp\| z \rpa\ \ewp\ewp^T +\\
    &+\int_0^{\infty} dx_1 \rho(x_1) \int_{-\infty}^{\infty} dz \rho(z)\ \theta\lpa \sqrt{d}\ma - w_1 x_1 - \|\w_\perp\| z \rpa\  \sum_{\beta} \int_{-\infty}^{\infty} dz_{\beta} \rho(z_\beta)\ z_\beta^2 \eb\eb^T +\\
    &+\int_0^{\infty} dx_1 \rho(x_1) \int_{-\infty}^{\infty} dz \rho(z)\ \theta\lpa \sqrt{d}\ma - w_1 x_1 + \|\w_\perp\| z \rpa\  \sum_{\beta} \int_{-\infty}^{\infty} dz_{\beta} \rho(z_\beta)\ z_\beta^2 \eb\eb^T =\\
    &= \int_0^{\infty} dx_1 \rho(x_1) \lsq 1+\erf\lpa\frac{\aw-\ampli x_1}{\sqrt{2}}\rpa \rsq \lpa  \ewp\ewp^T + \sum_{\beta} \eb\eb^T \rpa+\\
    &-\frac{2}{\sqrt{2\pi}} \int_0^{\infty} dx_1 \rho(x_1) e^{-\frac{\lpa\aw-\ampli x_1\rpa^2}{2}}\lpa\aw-\ampli x_1\rpa \ewp\ewp^T=\\
    &= \tilde{\I}_{d-1} \int_0^{\infty} dx_1 \rho(x_1) \lsq 1+\erf\lpa\frac{\aw-\ampli x_1}{\sqrt{2}}\rpa \rsq +\\
    &-\frac{2}{\sqrt{2\pi}} \int_0^{\infty} dx_1 \rho(x_1) e^{-\frac{\lpa\aw-\ampli x_1\rpa^2}{2}}\lpa\aw-\ampli x_1\rpa \ewp\ewp^T
\end{aligned}
\eeqn

where $\tilde{\I}_{d-1} = \I_d-\ewt\ewt^T$.

\end{document}